\documentclass[12pt]{elsarticle}

\usepackage{amsmath}
\usepackage{amsthm}
\usepackage{natbib}
\usepackage{amsfonts}
\usepackage{geometry}
\usepackage{graphicx}
\graphicspath{ {Figures/} }
\usepackage[outdir=./FiguresPDF/]{epstopdf} 
\usepackage{stmaryrd}
\usepackage{subcaption}
\usepackage{epstopdf}
\usepackage{amssymb}
\usepackage{upgreek}
\usepackage[left]{lineno}
\usepackage{hyperref}
\usepackage{changes}
\usepackage{lipsum}
\usepackage{textcomp}
\usepackage{soul}
\usepackage{slashbox}
\usepackage{algorithm}
\usepackage{algpseudocode}

\newcommand{\norm}[1]{\left\lVert#1\right\rVert}
\modulolinenumbers[1]
\biboptions{round,semicolon,sort&compress}

\journal{Elsevier}

\begin{document}

\begin{frontmatter}

\title{A physics-informed deep neural network for surrogate modeling in classical elasto-plasticity}

\author[ad1]{Mahdad Eghbalian\corref{cor1}}
\ead{meghbali@ucalgary.ca}

\author[ad2]{Mehdi Pouragha}
\ead{mehdi.pouragha@carleton.ca}

\author[ad1]{Richard Wan}
\ead{wan@ucalgary.ca}

\cortext[cor1]{Corresponding author}

\address[ad1]{University of Calgary, Dept. of Civil Engineering, Calgary, Canada}

\address[ad2]{Carleton University, Dept. of Civil and Environmental Engineering, Ottawa, Canada}

\begin{abstract}
In this work, we present a deep neural network architecture that can efficiently approximate classical elasto-plastic constitutive relations. The network is enriched with crucial physics aspects of classical elasto-plasticity, including additive decomposition of strains into elastic and plastic parts, and nonlinear incremental elasticity. This leads to a Physics-Informed Neural Network (PINN) surrogate model named here as Elasto-Plastic Neural Network (EPNN). Detailed analyses show that embedding these physics into the architecture of the neural network facilitates a more efficient training of the network with less training data, while also enhancing the extrapolation capability for loading regimes outside the training data. The architecture of EPNN is model and material-independent, i.e. it can be adapted to a wide range of elasto-plastic material types, including geomaterials and metals; and experimental data can potentially be directly used in training the network. To demonstrate the robustness of the proposed architecture, we adapt its general framework to the elasto-plastic behavior of sands. We use synthetic data generated from material point simulations based on a relatively advanced dilatancy-based constitutive model for granular materials to train the neural network. The superiority of EPNN over regular neural network architectures is explored through predicting unseen strain-controlled loading paths for sands with different initial densities.
\end{abstract}


\end{frontmatter}


\section{Introduction}
During the past few years, there has been a surge in using Machine Learning (ML) techniques, and especially Artificial Neural Networks (ANN), in non-STEM fields, such as social sciences and marketing, as well as different branches of STEM. Although the theoretical development of such techniques dates back a few decades, the main reasons behind their recent popularity is tied to the keyword ``Computational Power". On one hand, the immediate access to open-source GPU-based libraries, such as TensorFlow \citep{Abadi2016} and PyTorch \citep{NEURIPS2019_9015}, and machines with high computational power has increased; and on the other hand, algorithmic advancements, such as automatic differentiation and error back propagation, have resulted in lower computational power demand. The broader field of Mechanics has also seen a surge in using ML techniques, as the developments in ML presented the mechanics community with the opportunity to rely more on big data. This has led to the inception of a new sub-field in mechanics research known as Mechanistic Machine Learning.

\subsection{Data-driven Material Modeling}
Due to opportunities provided by Big Data Analytics, the field of material modeling has recently been faced with a change of paradigm from minimal use of data, followed in traditional symbolic model building, to maximal use of data in data-driven modeling. In material modeling, we deal with two categories of laws. First are the conservation laws, such as conservation of momentum and mass, and thermodynamic laws, which are universal laws based in general principles and have an epistemic nature. Second are the constitutive laws, such as stress-strain relationships in solids or velocity-pressure relationship in fluids, for which there is an epistemic uncertainty. The traditional mechanics resorts to symbolic forms for constitutive laws in an attempt to achieve reduction. However, reductionist approach comes at the cost of information loss, and this is precisely where developing modeling techniques that can rely more on data and potentially bypass the development of such symbolic models becomes appealing.

Two broad categories of approaches have been developed with the aim of replacing the traditional symbolic constitutive laws with data-driven models. First are the surrogate models based on ANNs (also known as parametric approaches), where the constitutive law is conceived in the form of a trained ANN. The second category pertains to the so-called ``Model-Free" approaches (also known as non-parametric approaches), where the constitutive law which acts as a surrogate to data is bypassed altogether and replaced with data. The basic idea behind the latter approach was first introduced in the seminal works of \citet{Kirchdoerfer2016, Kirchdoerfer2017} which was later on applied to different problems in mechanics, including elasticity \citep{Conti2018}, poroelasticity \citep{Bahmani2021}, inelasticity \citep{Eggersmann2019}, fracture mechanics \citep{Carrara2020}, and dynamics \citep{Kirchdoerfer2018}. A fundamental re-formulation of the basic governing equations is required within this approach whose computational handling requires developing numerical solvers with structures completely different from the common ones. While interesting, this latter approach is not explored in the current work.

Artificial Neural Networks can serve two general purposes for material modeling: (a) surrogate modeling and (b) parameter identification of constitutive laws with a priori assumed symbolic forms. In the latter case, the ANNs are used merely as a curve fitting tool for calibration (see e.g. \citet{Kotha2020}). Valuable information that might be hidden in the available data is potentially lost in this case due to the restrictions of the symbolic form of the model. Nevertheless, an interesting application of parameter identification using ANNs is presented in \cite{Flaschel2021} for discovery of hyperelastic constitutive laws from field data.

In the current work, we focus on the more exciting use of ANNs as a surrogate in constitutive modeling. In this case, the constitutive relation is not a priori known, but is intended to be discovered from the data. Provided that sufficient amount of data is available and general physical laws are appropriately hardwired into the ANN structure, the ANN can capture complex dependencies in the constitutive behavior of the material that was not otherwise possible in conventional model-based constitutive modeling \citep{Pouragha2020}. A summary of the state-of-the-art on using ANN as surrogate constitutive law, especially for elasto-plastic materials, is presented in the following.

\subsection{ANN as Surrogate Constitutive Law}
In their seminal work, \citet{Ghaboussi1991} proposed an ANN-based constitutive model for plane concrete and showed the capability of ANN in predicting the behavior of concrete under biaxial monotonic loading and uniaxial cyclic loading when presented with training data on the same loading paths. Their ANN was not general and was never tested for loading paths beyond the ones included in the training data. This work was followed by an abundance of research work on developing ANNs for elasto-plastic materials. \citet{Lefik2003} were one of the firsts to introduce a simple ANN for non-linear materials, including its implementation within a Finite Element framework. Their network was designed based on predicting the incremental behavior of the nonlinear material which was shown to perform superior to the one based on total behavior. A drawback of this work however was that the performed verification tests were not objective, as the same loading type used in training were also used for validation.

Plastic deformation of materials is a history-dependent process. In deep learning, such history-dependency is addressed via Sequence Learning, whose further elaboration leads to Recurrent Neural Networks (RNNs). For instance, \citet{Mozaffar2019} developed deep RNNs for modeling path-dependent plasticity with the time history of strain as input and the stress as output. The RNN was shown in this work to have superior capability over the regular ANN in predicting complex history-dependent plasticity; but its architecture is more complex, and it requires generating huge volume of data for training. In their recent works, \citet{Masi2021a, Masi2021b} showed that regular ANN architectures can be used for learning path-dependent behavior as long as a set of internal variables that represents the history of the materials (state variables) are properly included in the training of the ANN. In this work, we follow the same reasoning and avoid the complex structures in RNNs.

It is well-known that ANNs have poor extrapolation capability, i.e. they generally do not perform well for scenarios not included in the training data; a shortcoming that is rooted directly in their lack of proper physics bases \citep{Pouragha2020}. In the case of constitutive behavior, this can have disastrous consequences in practice if the ANN is used in a boundary value problem setting to study extreme events, such as failure of bolts in airplane wings or embankment failure, where it can be subjected to complex loading paths never included in the training data. The poor extrapolation capability of ANNs has led to the development of Physics-Informed Neural Networks (PINNs) where the ANNs are endowed with some of the physics relevant to the phenomenon being modeled. The aim is to reduce the possibility of having non-physical outputs by enforcing physics-based constraints, and thereby improving the stability of the models when implemented for a particular boundary value problem. PINNs were primarily introduced for the Inversion problems \citep{Raissi2019,Haghighat2021}, but have recently permeated also in surrogate modeling of constitutive laws.

As the use of PINNs as surrogate constitutive models is relatively new, only a few noteworthy works can be mentioned. \citet{Liang2008} presented a PINN for surrogate modeling of hyper-elastic foams. Their network learns the strain energy potential function directly from data following Sobolev training. The stress is then obtained based on calculating the derivatives of the output. The PINN in this work is mainly used for curve fitting the data without the attempt to investigate the capability of the network to interpolate or extrapolate the behavior of foams. \citet{Masi2021a, Masi2021b} presented a PINN for constitutive modeling of plastic materials that follow the thermodynamics consistent hyper-plasticity theory. The network predicts the hyper-plasticity potential function following Sobolev training, from which the state variables are derived via differentiation. They showed the superiority of their Thermodynamics-informed ANN (TANN) to the regular ANN in terms of data efficiency and generalization capability. Their network provides excellent predictions for loading paths never seen in training. However, in the case where the potential function (output of the network) is not known, as is the case with experimental data, the training of the network based only on the derivatives of the network becomes challenging and may give rise to non-unique network parameters.

\citet{Vlassis2021, Vlassis2022} presented an ANN-based framework for conventional elasto-plastic materials that incorporates a yield surface and plastic flow rule. The yield surface is recast as a level-set function that can evolve in the stress space (representing hardening/softening) following a prescribed rule. The network is trained using Sobolev training as their cost functions involve the derivative of the network outputs. Their ANN provides crucial access to interpretable plasticity laws directly from data. However, data regarding the initial position of the yield surface must be presented to the network which may not be practical when using real experimental data for training. Moreover, the evolution of the yield surface follows a priori assumed law which might be restrictive. In another work, \citet{Xu2021} presented a PINN for inelastic materials obeying classical elasto-plasticity, hypo-plasticity and hyper-plasticity. The PINN is trained based on predicting the Cholesky factor of the tangent elasto-plastic matrix. The full tangent matrix can then be re-constructed from the output of the network. This results in a tangent matrix which is symmetric positive definite, leading to satisfying the second-order work criterion \citep{Nicot2017}. Such network, however, is more useful in an academic setting as the tangent elasto-plastic matrix required for training is only available in synthetic data generated from known models and not in practice from experimental data. 

In the current work, we focus on conventional elasto-plasticity and develop a PINN, named here as Elasto-Plastic Neural Network (EPNN), with simple architecture, that incorporates some of the crucial physics related to this class of plastic materials, and is tailored for use with experimental data. The proposed EPNN replaces the conventional yield function, plastic potential, and plastic flow rule with more flexible algorithms. We use synthetic data for training and verification of our network which are generated from a relatively advanced model for sands \citep{Wan1998,Wan1999}. The comparisons with regular ANNs indicate that embedding the relevant physics into the EPNN significantly improves its generality in capturing unseen loading paths. The robustness of the model is also shown to improve, as the possibility of having progressively accumulating errors is diminished.

\subsection{Organization and Notations}

The organization of the paper is as follows: Section \ref{sec:background} provides some background on the mathematical formulation and structure of ANNs as well as the general formulation of conventional elasto-plasticity; Section \ref{sec:ann-general} presents the general architecture of EPNN proposed in this work; Sections \ref{sec:WG} and \ref{sec:verification} demonstrate the training performance and advantages of this newly developed architecture over regular ANNs in a case study; and finally, the conclusions and outlook on future research on this topic are presented in Section \ref{sec:conclusions-and-outlook}.

In this work, we use boldface symbols (for instance $\mathbf{a}$) to represent vectors and tensors, while regular symbols (for instance $a$) denote scalars. Calligraphic symbols, such as $\mathcal{Z}$, are used to represent sets. Regular symbols with subscript are used to represent the components of vectors and tensors. Subscripts are typeset in italics and repetition of indices follows Einstein's summation convention. Both boldface and regular symbols can have superscripts which are typeset in text mode and do not follow Einstein's convention. Their meaning will be explained upon definition.

\section{Background}\label{sec:background}

\subsection{Artificial Neural Networks Overview}\label{sec:background-ann}
In this section, the internal mechanism of ANNs is briefly recalled for readers of general engineering background. Readers familiar with basics of ANNs can safely skip this part. 

\begin{figure}[t]
	\begin{center}
		\includegraphics[width=0.7\linewidth]{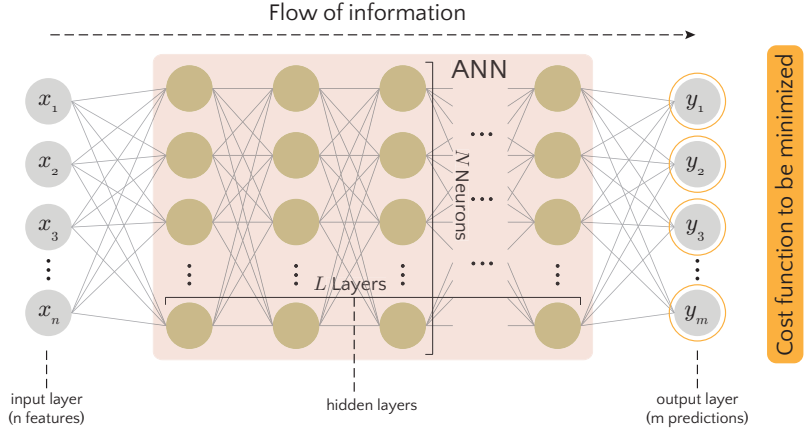}
		\caption{\label{fig:1} Structure of a simple ANN. Orange circles signify the variables used for calculation of the cost function (filled orange box).}
	\end{center}
\end{figure}

ANN is a popular ML algorithm with wide range of applications as a universal approximator of continuous highly nonlinear functions of many variables. ANN learns from data presented to it using a mathematical model that simulates the function of neurons in the brain. A simple ANN structure is shown in Fig. \ref{fig:1}. A set of simple processing units called ``neurons" are defined in ANN that are grouped into layers. The layers of an ANN are categorized as input layer, hidden layers and output layer. The input layer includes the input ``features" to the network, while the output layer includes the output predictions of the network.

In ANN, the information flows through neurons which are connected by ``links" (gray lines in the network of Fig. \ref{fig:1}). The ``strength" of each link in ANN is determined by a variable weight. The weights of all the links between two adjacent layers $\text{i}-1$ and $\text{i}$ are shown by a matrix $\mathbf{W}^{\text{i}-1,\text{i}}$ where each component $W_{kl}^{\text{i}-1,\text{i}}$ represent the weight between $k^{\text{th}}$ neuron in layer $\text{i}-1$ and $l^{\text{th}}$ neuron in layer $\text{i}$. The input to the $l^{\text{th}}$ neuron in layer $\text{i}$ ($z_l^{\text{i}}$) is then defined as the linear combination of the outputs (shown by symbol $a$) of the connected neurons in layer $\text{i}-1$ as:
\begin{equation}\label{eq:4}
	z_l^{\text{i}} = W_{kl}^{\text{i}-1,\text{i}} a_k^{\text{i}-1} + b^{\text{i}-1}, \quad k = 1:N^{\text{i}-1} \quad \quad \text{or} \quad \quad \mathbf{z}^{\text{i}} = \mathbf{a}^{\text{i}-1} \mathbf{W}^{\text{i}-1,\text{i}} + b^{\text{i}-1} \mathbf{I},
\end{equation}
where $b^{\text{i}-1}$ is the bias term, $N^{\text{i}-1}$ is the number of neurons in layer $\text{i}-1$, and $\mathbf{I}$ refers to the unity vector with all elements being $1$. The output of the neuron is then determined by activating the input using a nonlinear activation function $g$:
\begin{equation}\label{eq:6}
	a_l^{\text{i}} = g\left(z_l^{\text{i}}\right).
\end{equation}
Normally, no activation function is used for the neurons of the output layer. The activation function of the neurons is the source of nonlinearity in the ANN and gives them the ability to learn highly nonlinear mappings. Given a set of known input-output (or features-labels) pairs, the input is transmitted through the network and an output (prediction) is calculated. The error is then computed based on the mismatch between the network output and the known labels. The ``training" or ``learning" of ANN is the process of finding optimal values for its link weights ($W_{kl}^{\text{i}-1,\text{i}}$) such that this error is reduced below a desirable accuracy. This goal can be mathematically formulated as a minimization problem over a cost function. The cost function in Fig. \ref{fig:1} is shown inside a filled orange box. Throughout this paper, colored circles shown in the structure of neural networks signify the variables used for calculation of the filled box with the same color. In this work, we use the Mean Squared Error (MSE) and Mean Absolute Error (MAE) cost functions. The MSE cost function reads:
\begin{equation}
	\text{MSE}^{\text{network}} = \frac{1}{M} \sum_{\text{i}=1}^{M} \text{MSE}^{\text{i}} \quad \text{where} \quad \text{MSE}^{\text{i}}=\norm{\mathbf{y}^{\text{i}} - \overline{\mathbf{y}}^{\text{i}}}^2,
\end{equation}
where $M$ is the size of training data, $\mathbf{y}^{\text{i}}$ includes the predictions of the network based on the input features of the $\text{i}^{\text{th}}$ training data, and $\overline{\mathbf{y}}^{\text{i}}$ is the label of the $\text{i}^{\text{th}}$ training data. The MAE cost function on the other hand reads:
\begin{equation}
	\text{MAE}^{\text{network}} = \frac{1}{M} \sum_{\text{i}=1}^{M} \text{MAE}^{\text{i}} \quad \text{where} \quad \text{MAE}^{\text{i}}=\sum_{j=1}^{m} | y^{\text{i}}_j - \overline{y}^{\text{i}}_j |,
\end{equation}
where $y^{\text{i}}_j$ is the $j^{\text{th}}$ output (prediction) of the network based on the input features of the $\text{i}^{\text{th}}$ training data, and $\overline{y}^{\text{i}}_j$ is the $j^{\text{th}}$ label of the $\text{i}^{\text{th}}$ training data.

Several optimization algorithms exist for minimizing the cost function. In this work, we use Adaptive Moment Estimation (ADAM) optimizer which, like many other algorithms, involves a ``learning rate" that determines the step size taken in each epoch for correction of the network weights.

Finally, it should be mentioned that the performance of ANN is highly dependent on the quality and quantity (size) of the data used for its training. The quality of the data is related to how representative it is of the process to be simulated by ANN. It is only through presenting ANN with a high quality data that we can expect it to predict new results. The argument presented in \citet{Pouragha2020} concluded that, if a constitutive model exists for a material, then in principle, such a training dataset with a finite size also exists from which the behavior of materials can be comprehensively learned.

\subsection{Problem Setup}

\subsubsection{Governing Field Equations of Solid Bodies}
Within the general framework of continuum mechanics, the set of governing field equations for a solid body undergoing small strain in the static case reads:
\begin{subequations}\label{eq:3}
	\begin{alignat}{4}
		&\dot{\sigma}_{ij}(\mathbf{x})=\Sigma_{ij}(\sigma_{ij}(\mathbf{x}), \mathcal{I}(\mathbf{x}), \varepsilon_{ij}(\mathbf{x}), \dot{\varepsilon}_{ij}(\mathbf{x})) \quad\forall \mathbf{x}\in \Omega, \label{eq:3a} \\
		&\frac{\partial \sigma_{ij}(\mathbf{x})}{\partial x_j} + b_i(\mathbf{x})=0\quad\forall \mathbf{x} \in \Omega, \label{eq:3b} \\
		&\varepsilon_{ij}(\mathbf{x})=\frac{1}{2}\left(\frac{\partial u_i(\mathbf{x})}{\partial x_j}+\frac{\partial u_j(\mathbf{x})}{\partial x_i}\right)\quad\forall \mathbf{x} \in \Omega, \label{eq:3c}\\
		& u_i(\mathbf{x})=\overline{u}_i(\mathbf{x}) \quad\forall \mathbf{x} \in \Gamma^\text{u}, \label{eq:3d}\\
		& \sigma_{ij}(\mathbf{x})n_j(\mathbf{x})=\overline{t}_i(\mathbf{x}) \quad \forall \mathbf{x} \in \Gamma^\text{t}. \label{eq:3e}
	\end{alignat}
\end{subequations}

In Eq. \eqref{eq:3}, $\mathbf{x}$ is the position vector; $\Omega$ and $\Gamma$ are the physical domain of the solid body and its boundary, respectively; $\sigma_{ij}(\mathbf{x})$, $\varepsilon_{ij}(\mathbf{x})$, $ u_i(\mathbf{x})$ and $b_i(\mathbf{x})$ are the Cauchy stress, strain, displacement and body force fields, respectively; and $\mathcal{I}$ represents the set of internal state variables such as plastic strain. Also, the superimposed dot refers to increment. In the boundary conditions (Eqs. \eqref{eq:3d} and \eqref{eq:3e}), $\Gamma^\text{u}$ is the part of boundary with prescribed displacement $\overline{u}_i$, while $\Gamma^\text{t}$ is the part of boundary with prescribed stress traction $\overline{t}_i$. Finally, Eq. \eqref{eq:3a} is the constitutive relation. In the following section, more details are provided on the general form of constitutive relations for an inelastic material that follows the conventional elasto-plasticity.

\subsubsection{Features of Elasto-Plastic Models}\label{sec:const}
This section presents the basic ingredients and general form of constitutive relations for a class of plasticity models that follow the classical continuum theory of plasticity \citep{Hill1959,Hill1962}. Conventional plasticity is predicated on a few postulates which will be briefly reviewed in the following.

We restrict ourselves here to the case of small deformations. The first fundamental postulate of classical plasticity is the additive decomposition of total strain into elastic and plastic components: $\varepsilon_{ij}=\varepsilon^\text{e}_{ij} + \varepsilon^\text{p}_{ij}$ where $\varepsilon_{ij}$, $\varepsilon^e_{ij}$ and $\varepsilon^\text{p}_{ij}$ refer to total strain, elastic strain and plastic strain, respectively. The elastic strain can be specified through hypo-elasticity, general elasticity or hyper-elasticity. Irrespective of the elasticity model used, the rate of the elastic strain $\dot{\varepsilon}^\text{e}_{ij}$ can be related to the rate of stress $\dot{\sigma}^\text{e}_{ij}$ through a constitutive matrix (tangent elasticity tensor) $C_{ijkl}$:
\begin{equation}\label{eq:1}
	\dot{\sigma}_{ij}=C_{ijkl}\dot{\varepsilon}^\text{e}_{kl}.
\end{equation}

In case of hypo-elasticity, $C_{ijkl}$ is a fourth-order tensor function of elastic strain; in case of general elasticity, it is the derivative of a second-order tensor function of elastic strain, with respect to elastic strain; and finally in case of hyper-elasticity, it is the second-order derivative of a scalar function of elastic strain, with respect to elastic strain. Thus, general elasticity and hyper-elasticity are considered as special cases of the more general hypo-elasticity.

The plastic strain, on the other hand, is determined based on the concept of yield surface and the flow rule postulate. The following set of equations describe conventional elasto-plasticity:
\begin{subequations}\label{eq:2}
	\begin{alignat}{4}
		&\dot{\sigma}_{ij}=C_{ijkl}\left(\dot{\varepsilon}_{kl}-\dot{\varepsilon}^\text{p}_{kl}\right), \label{eq:2a}\\
		&F(\sigma_{ij},\mathcal{H}) \le 0, \label{eq:2b}\\
		&\dot{\varepsilon}^\text{p}_{ij}= \dot{\lambda}\frac{\partial P(\sigma_{ij},\mathcal{H})}{\partial \sigma_{ij}}, \label{eq:2c}\\
		&\dot{\lambda} \ge 0,~~F \le 0,~~\dot{\lambda} F=0, \label{eq:2d}\\
		&\dot{\mathcal{H}} = H(\mathcal{H},\sigma_{ij}, \varepsilon_{ij},\mathcal{I},\dot{\sigma}_{ij}, \dot{\varepsilon}_{ij},\dot{\mathcal{I}}), \label{eq:2e}
	\end{alignat}
\end{subequations}
where $F(\sigma_{ij},\mathcal{H})$ is the yield function describing the boundary between elastic and plastic zones in the stress space; Eq. \eqref{eq:2c} is the flow rule written based on a plastic potential $P(\sigma_{ij},\mathcal{H})$ and the plastic multiplier $\dot{\lambda}$ which determines the magnitude of the plastic strain increment; and Eq. \eqref{eq:2d} is the classical Melan-Kuhn-Tucker relation. Moreover, $\mathcal{H}$ signifies the set of hardening variables which are generally functions of stress, strain and internal state variables, mainly the plastic strain.

Formulation of the yield and plastic potential functions is often difficult for materials with complex plastic behavior, and various complex forms have been introduced to achieve an accurate description of the plastic behavior of different materials. We aim here at developing a proper PINN embedded with salient physics of elasto-plasticity which enables us to describe the behavior of elasto-plastic materials, without having to symbolically construct equations for elasticity, yield function, and plastic potential.

It should be re-iterated here that the discussion in this section pertains to the classical theory of plasticity. However, there are other models for determining the plastic strain based on hypo-plasticity, such as the incrementally non-linear formulation of \citet{Darve1995}, and hyper-plasticity \citep{Houlsby2007}.

\section{ANN-based Elasto-plastic Constitutive Model}\label{sec:ann-general}
The general forms of elasto-plasticity constitutive equations were presented in Section \ref{sec:const} (Eq. \eqref{eq:2}). The set of equations can be seen as a highly nonlinear mapping between a set of inputs and outputs. Such mapping can be instead represented by an ANN consisting of a similar set of input-output pairs. This section presents the general architecture of two simple ANNs that can replace the set of elasto-plastic constitutive equations. Next, a novel physics-informed ANN (PINN), herein called EPNN, will be presented and its advantages over the regular ANNs will be demonstrated in Section \ref{sec:verification}. The ground truth for evaluating the accuracy of the EPNN and ANNs is obtained through numerically integrating the system of equation in Eq. \eqref{eq:2}.

\subsection{Regular ANN Architecture}\label{sec:ANN-general}
This section presents two regular ANN architectures for representing elasto-plasticity which will be used as baselines to evaluate the performance of the eventual EPNN presented in Section \ref{sec:PINN-general}. The procedure followed in numerical integration of system of Eqs. \ref{eq:2} (implicit or explicit) in a strain-controlled step involves calculating stress, strain and internal state variables at step $\text{n}+1$ $\{\sigma^{\text{n}+1}_{ij}, \varepsilon^{\text{n}+1}_{ij}, \mathcal{I}^{\text{n}+1}\}$ (or their increments), based on their values at step $\text{n}$, and the known strain increment $\Delta\varepsilon^{\text{n}}_{ij}$. As such, the input to ANN should include $\{\sigma^{\text{n}}_{ij}, \varepsilon^{\text{n}}_{ij}, \mathcal{I}^{\text{n}}, \Delta\varepsilon^{\text{n}}_{ij}\}$, while the network should predict the changes in stress and internal state variables $\{\Delta\sigma^{\text{n}}_{ij}, \Delta\mathcal{I}^{\text{n}}\}$ as its output.
\begin{figure}
	\begin{center}
		\includegraphics[width=0.99\linewidth]{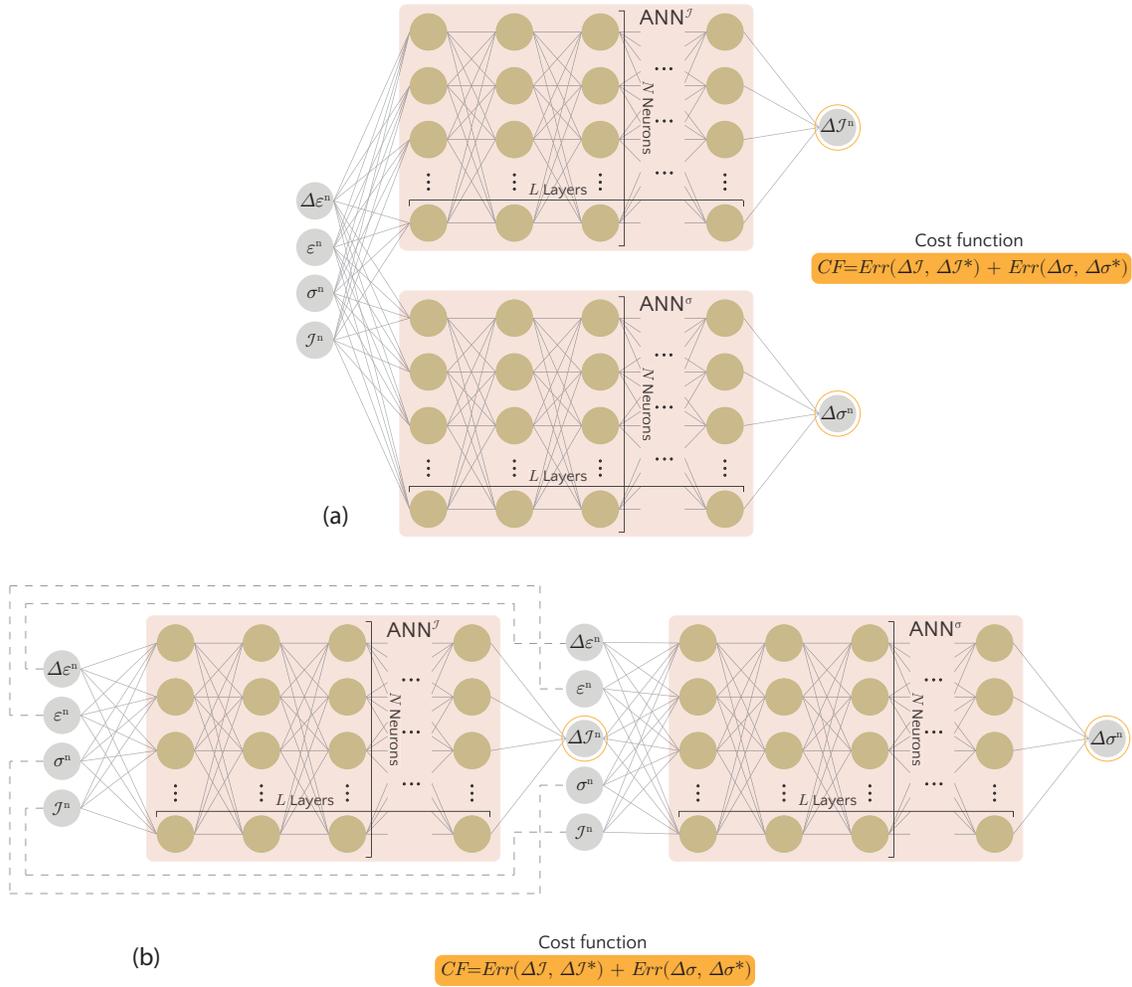}
		\caption{\label{fig:2} (a) Parallel and (b) serial ANN architectures used in this work for comparison with the newly proposed EPNN in Section \ref{sec:PINN-general}.}
	\end{center}
\end{figure}

When representing elastoplastic materials with a regular ANN, two different architectures can be considered: ``parallel" architecture and ``serial" architecture (see Fig. \ref{fig:2}). The serial ANN architecture was originally proposed by \citet{Masi2021b} to validate their TANN, while the parallel architecture is a simpler version used in this work for comparison. The former serves as the basis for constructing the EPNN in Section \ref{sec:PINN-general}. In the parallel architecture, the increment of stress and internal state variables are predicted based on two separate sub-networks (internal state sub-network $\text{ANN}^{\mathcal{I}}$ and stress sub-network $\text{ANN}^{\upsigma}$). Since there are no dependencies between the two sub-networks, they can be trained separately or simultaneously. Simultaneous training of the sub-networks can be done by defining a total cost function $CF$ as the sum of cost functions of the two sub-networks, i.e. $CF=CF^\mathcal{I}+CF^{\upsigma}$, where:
\begin{subequations}\label{eq:16}
	\begin{alignat}{2}
		CF^\mathcal{I} &=Err(\Delta\mathcal{I}, \Delta\mathcal{I}^*), \label{eq:16a}\\
		CF^{\upsigma} &=Err(\Delta \boldsymbol{\sigma}, \Delta \boldsymbol{\sigma}^*). \label{eq:16b}
	\end{alignat}
\end{subequations}

In Eq. \eqref{eq:16}, $Err$ is the error (cost) function (for instance MAE or MSE); $\Delta\mathcal{I}$ and $\Delta \boldsymbol{\sigma}$ are the outputs of internal state and stress sub-networks, respectively; and $\Delta\mathcal{I}^*$ and $\Delta \boldsymbol{\sigma}^*$ are the labeled data.

In the serial architecture, on the other hand, the internal state sub-network first predicts the increments of internal state variables which are then used as additional inputs to the stress sub-network. As such, the two sub-networks cannot be trained separately. The cost function of the serial network has the same form as the parallel network (Eq. \eqref{eq:16}); however, the training behaviors of the two networks are different, particularly because, in the serial architecture, the weights of the internal state sub-network also influence the cost function of the stress sub-network.

It is seen in the architectures shown in Fig. \ref{fig:2} that there is no trace of strain decomposition, elasticity, yield function, plastic potential, flow rule and hardening laws in the ANN-based elasto-plasticity models. The ANN serves as a mere general mapping where all these features are bypassed. Bypassing these physics leads the ANN model to be more data-hungry for calibration (learning) and weaker in generalization of the material behavior compared to the model-based approach.

In Sections \ref{sec:WG} and \ref{sec:verification}, the training performance and predictions of these two ANN architectures will be compared with EPNN in a case study.

\subsection{EPNN Architecture}\label{sec:PINN-general}
In this section, we propose a novel PINN architecture by enriching the serial ANN architecture presented in Section \ref{sec:ANN-general} with crucial physics related to the elasto-plasticity framework explained in Section \ref{sec:const}. The physics we incorporate in the serial ANN architecture include: the additive decomposition of strain into elastic and plastic parts, and hypo-elasticity. Recalling Eq. \eqref{eq:2a}, the stress-strain relation can be written in incremental form as:
\begin{equation}\label{eq:12}
	\Delta \sigma_{ij} = C^{\text{sec}}_{ijkl} \left(\Delta \varepsilon_{kl} - \Delta \varepsilon^\text{p}_{kl}\right),
\end{equation}
where the secant elasticity tensor $C^{\text{sec}}_{ijkl}$ is generally a function of stress, strain, internal state variables and their increments.

\begin{figure}
	\begin{center}
		\includegraphics[width=0.98\linewidth]{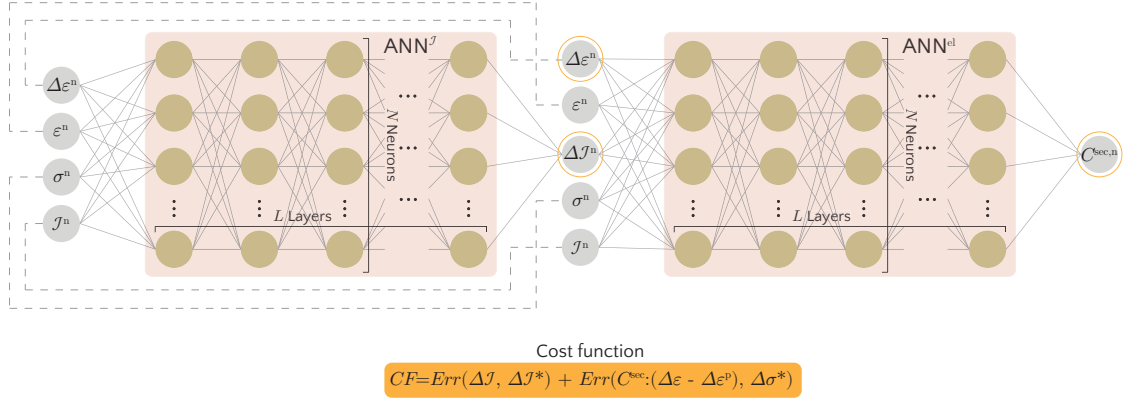}
		\caption{\label{fig:3} The general EPNN architecture proposed in this work for materials following classical elasto-plasticity.}
	\end{center}
\end{figure}

Figure \ref{fig:3} shows the proposed EPNN architecture. Compared to the serial ANN architecture in Section \ref{sec:ANN-general}, the stress sub-network is replaced by an elasticity sub-network which outputs the secant elasticity tensor of the material. As such, given the elasticity tensor (output of the elasticity sub-network), strain increment (input of EPNN) and plastic strain increment (output of the internal state sub-network), we can calculate the stress increment using Eq. \eqref{eq:12}. It should be noted that the plastic strain is included in the internal state set, $\mathcal{I}$. In contrast to the serial ANN, here, the stress increment is an indirect output of the elasticity sub-network which follows the hyper-elasticity and strain decomposition assumptions. The cost function of EPNN is thus defined as $CF=CF^\mathcal{I}+CF^{\upsigma}$ where $CF^\mathcal{I}$ has the same form as in Eq. \eqref{eq:16a}, while $CF^{\upsigma}$ reads:
\begin{equation}\label{eq:17}
	CF^{\upsigma} =Err(\mathbf{C}^{\text{sec}} : \left(\Delta \boldsymbol{\varepsilon}- \Delta \boldsymbol{\varepsilon}^{\text{p}}\right), \Delta \boldsymbol{\sigma}^*).
\end{equation}
Therefore, EPNN employs a modified cost function which incorporates some physics of elasto-plasticity. It is important to mention that, since the elasticity sub-network outputs the elasticity matrix, the network requires the labels of the elasticity matrix for efficient training. However, in this work we train the EPNN without providing it with this information. This comes at the cost of less training efficiency, i.e. slower convergence to the optimal solution. Training the EPNN without the labels of the elasticity matrix is done for two reasons: (1) to use the exact same set of inputs and labels used for training the regular ANNs in Section \ref{sec:ANN-general}, without requiring any additional information, and (2) such information (elasticity matrix) is normally not available when using experimental results for training the network.

\section{Application to Plasticity of Sands: Model Training}\label{sec:WG}
In this section, we show the application of the general neural network-based framework for elasto-plastic materials introduced in Section \ref{sec:ann-general} in a case study. We use synthetic data as our ground truth to first study different aspects of training a neural network-based constitutive model, and secondly to investigate the robustness of the newly proposed EPNN architecture in this work and the advantages it brings with respect to regular ANN architectures.

We use the constitutive model for sands developed in \citep{Wan1998,Wan1999} to generate the synthetic data. The model, named here as WG model, is capable of describing many salient characteristics of sands observed in experiments, such as dependency on confining pressure and void ratio; thus, the synthetic data generated from it naturally encapsulates these complex characteristics. \citet{Wan2015} successfully used this model to investigate strain localization and different precursors to failure in sands. The basic ingredients of this model are briefly summarized in \ref{appendix:1}. Details about synthetic data generation and post-processing, the network architectures adapted to sands, choice of network parameters, and learning and training curves are presented in the subsequent parts of this section.

\subsection{Data Generation and Post-processing}
Overall, the WG model includes 10 calibration parameters: $\varphi^{\text{cs}}$, $\beta$, $a$, $e^{\text{cs},0}$, $h$, $n$, $\mu$, $G^0$, $\nu$, $\alpha$ which are defined in \ref{appendix:1}. These parameters have been calibrated for different sands in \citet{Pinheiro2009} (Chapter 3). We choose Ottawa sand for the generation of synthetic data. The parameters of the WG model calibrated for this sand are presented in \citet{Pinheiro2009} (Chapter 3, Table 3-4) and are also repeated here in Table \ref{Table:1}.

\begin{table}[h]
	\small
	\renewcommand{\arraystretch}{1.2}
	\caption{Material parameters of the WG model for Ottawa sand.}
	\centering
	\begin{tabular}{c c c c c c c c c c}
		\hline
		$G^0$ (kPa) & $\nu$ & $\mu$ & $\sin{\varphi^{\text{cs}}}$ & $\beta$ & $a$ & $e^{\text{cs},0}$ & $h$ (kPa) & $n$ & $\alpha$ \\ \hline
		$9.0\mathrm{e}{+02}$ & $0.3$ & $0.8$ & $0.53$ & $1.3$ & $8.0\mathrm{e}{-03}$ & $0.74$ & $5.65\mathrm{e}{5}$ & $0.4$ & $1.5$ \\
		\hline
	\end{tabular}
	\label{Table:1}
	\renewcommand{\arraystretch}{1}
\end{table}
 
Numerical integration of the WG model is done implicitly. To reduce the size of training data, stress integration was done in the principal stress/strain space. The Python source code of the WG model is accessible through the following repository:
\begin{equation*}
	\text{\href{https://github.com/meghbali/WGImplicit}{https://github.com/meghbali/WGImplicit}}
\end{equation*}

For modeling the sand behavior, the WG model requires two initial conditions defined for the numerical sample: initial confining stress $p^{\text{in}}$, and initial void ratio $e^{\text{in}}$. To sweep the possible ranges of initial conditions, we perform virtual tests on a $10 \times 10$ grid of initial conditions in the range $p^{\text{in}} \in [50.0, 500.0]$ kPa and $e^{\text{in}} \in [0.5, 0.74]$. For each initial condition, we perform $20$ monotonic strain-controlled tests in the 3D true triaxial principal strain/stress space, by applying proportional strain paths in terms of $\{\varepsilon_{11}, \varepsilon_{22}, \varepsilon_{33}\}$. The tests are performed in a maximum of $200$ steps. The direction of strain loading is chosen randomly for each of the $20$ tests and is fixed in all the steps, but its magnitude in each step is randomly selected from the range $[0.0, 1.6\mathrm{e}{-03}]$ to ensure first-order homogeneity of the incremental stress-strain relation. The time step pairs are used as the set of input-label pairs for training the neural networks. A total of $\approx 310,000$ samples are generated for training. This dataset can be found at \href{https://github.com/meghbali/ANNElastoplasticity}{https://github.com/meghbali/ANNElastoplasticity}.

\begin{figure}
	\begin{center}
		\includegraphics[width=0.99\linewidth]{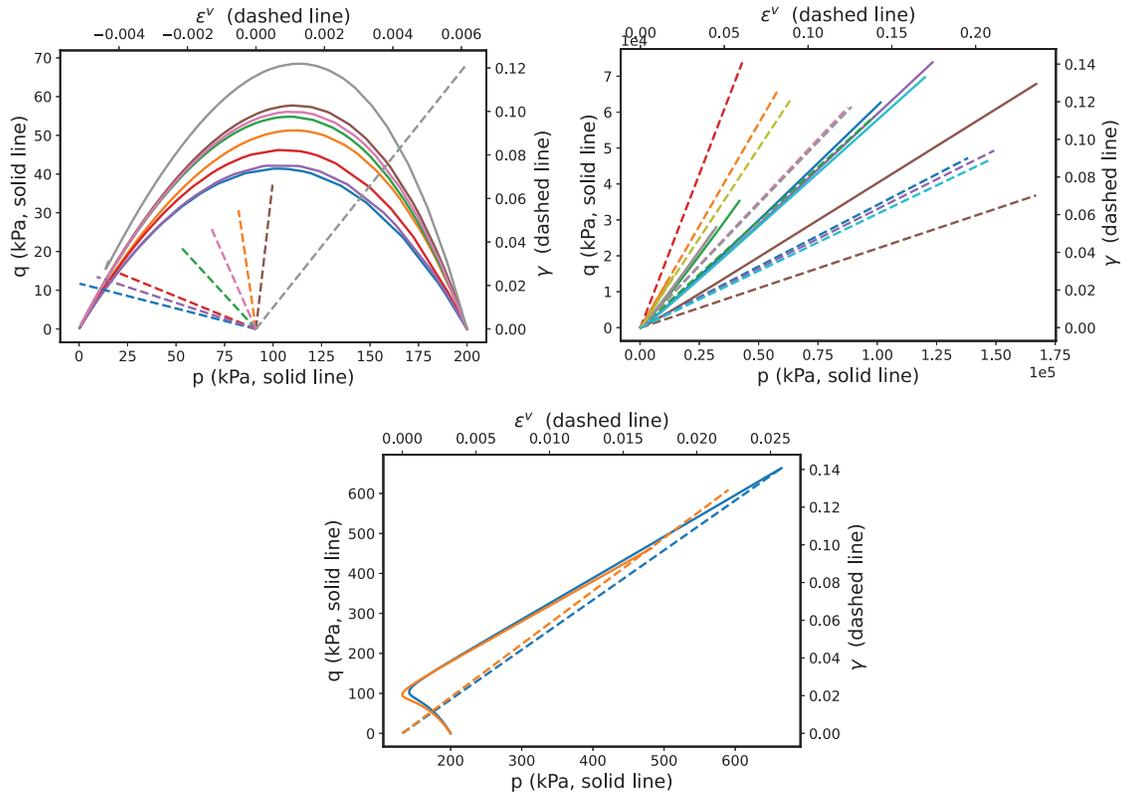}
		\caption{\label{fig:4} Samples of random monotonic strain path (and the ensuing stress paths) followed in generating the dataset used for training. All the paths corresponding to $p^{\text{in}}=200$ kPa and $e=0.74$ are shown here. The dashed lines refer to the strain path (volumetric strain $\varepsilon^{\text{v}}$ versus equivalent shear strain $\gamma$), while the solid lines with the same color show the corresponding stress paths (mean stress $p$ versus equivalent shear stress $q$).}
	\end{center}
\end{figure}

Samples of the random monotonic loading paths followed for generating the dataset are presented in Fig. \ref{fig:4}.
\begin{figure}
	\begin{center}
		\includegraphics[width=1.0\linewidth]{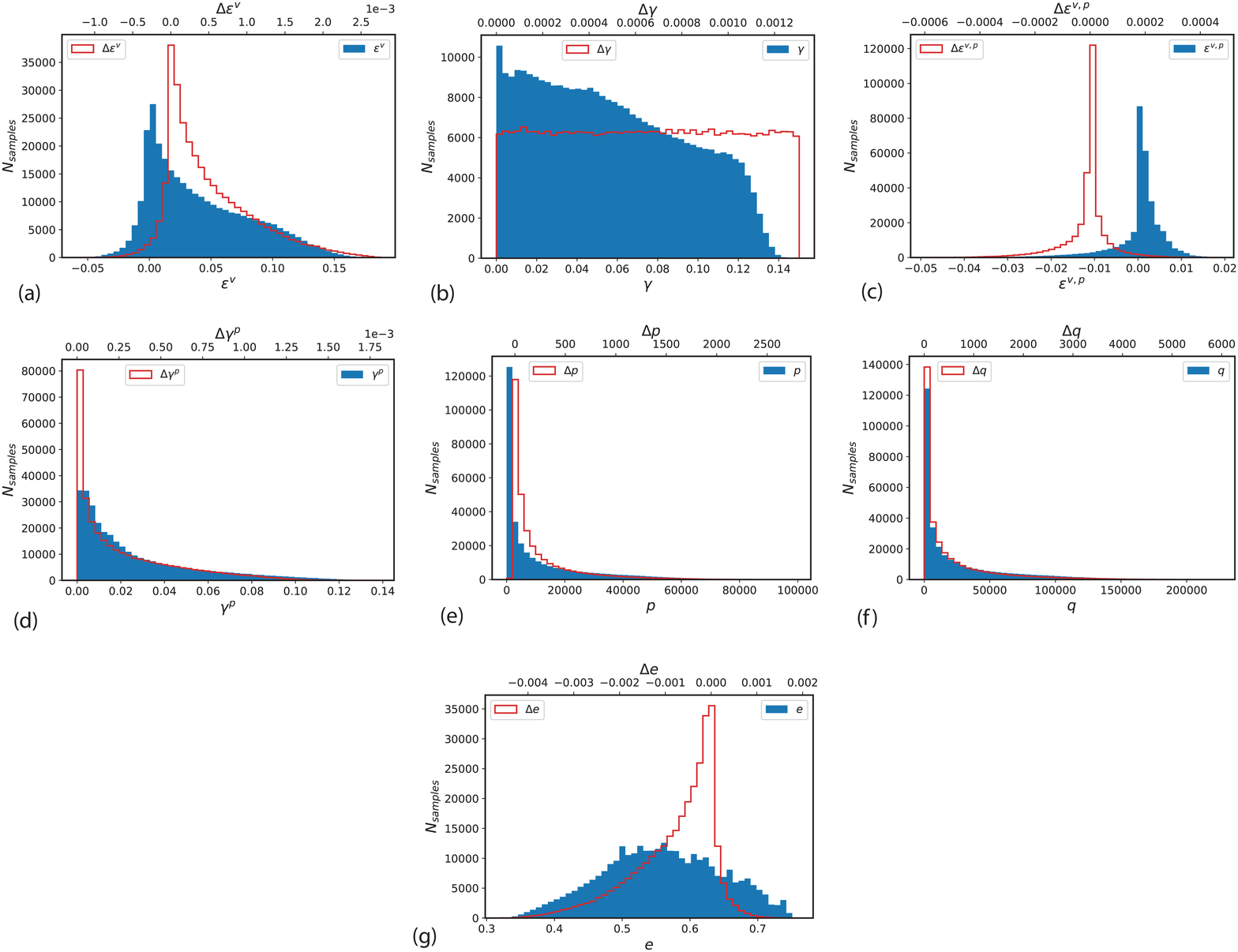}
		\caption{\label{fig:34} Sampling for the dataset used for training the networks in this work: (a) volumetric strain, (b) equivalent shear strain, (c) volumetric plastic strain, (d) equivalent plastic shear strain, (e) mean stress, (f) equivalent shear stress, and (g) void ratio.}
	\end{center}
\end{figure}
It should be re-iterated that each step (increment of loading) in the proportional loading path is treated as a separate data sample for training of the neural networks. The dataset sampling is shown in Fig. \ref{fig:34}, where for simplicity, the 3D stress and strains are represented via their mean and shear invariants. As can be seen in Figs. \ref{fig:34}a and \ref{fig:34}b, because of adopting a random magnitude at each loading step of the proportional strain path, the volumetric and shear strain increments include a wide range of values which helps the neural network be less sensitive to the step size when used in recall mode. Figures \ref{fig:34}e and \ref{fig:34}f also show the wide range of mean and shear stresses included in the dataset. The reason for seeing large number of samples having mean and shear stresses close to zero is that a small number of the loading paths followed in generating the dataset produce stresses that are orders of magnitudes larger than the rest of the paths. Finally, Fig. \ref{fig:34}g shows that the dataset includes void ratios that cover a wide range of sand densities from very loose ($e^{\text{in}}\approxeq 0.3$) to very dense ($e^{\text{in}}\approxeq 0.75$). Some statistical information related to distributions shown in Fig. \ref{fig:34} is presented in Table \ref{Table:4}.
\begin{table}
	\scriptsize
	\renewcommand{\arraystretch}{1.2}
	\caption{Mean, minimum (min), maximum (max), and standard deviation (sd) values for the distributions shown in Fig. \ref{fig:34}.}
	\centering
	\begin{tabular}{c c c c c c}
		\hline
		Data & unit & mean & min & max & sd \\ \hline
		$\varepsilon^{\text{v}}$ & - & $4.05\mathrm{e}{-02}$ & $-5.86\mathrm{e}{-02}$ & $1.87\mathrm{e}{-01}$ & $4.1\mathrm{e}{-02}$ \\
		$\Delta \varepsilon^{\text{v}}$ & - & $5.1\mathrm{e}{-04}$ & $-1.2\mathrm{e}{-03}$ & $2.7\mathrm{e}{-03}$ & $5.7\mathrm{e}{-04}$ \\
		$\gamma$ & - & $5.7\mathrm{e}{-02}$ & $0.0$ & $1.5\mathrm{e}{-01}$ & $3.7\mathrm{e}{-02}$ \\
		$\Delta \gamma$ & - & $6.5\mathrm{e}{-04}$ & $1.9\mathrm{e}{-08}$ & $1.3\mathrm{e}{-03}$ & $3.8\mathrm{e}{-04}$ \\
		$\varepsilon^{\text{v},\text{p}}$ & - & $6.9\mathrm{e}{-05}$ & $-4.9\mathrm{e}{-02}$ & $1.9\mathrm{e}{-02}$ & $6.6\mathrm{e}{-03}$ \\
		$\Delta \varepsilon^{\text{v},\text{p}}$ & - & $-1.2\mathrm{e}{-05}$ & $-6.0\mathrm{e}{-04}$ & $4.7\mathrm{e}{-04}$ & $8.4\mathrm{e}{-05}$ \\
		$\gamma^{\text{p}}$ & - & $2.7\mathrm{e}{-02}$ & $0.0$ & $1.4\mathrm{e}{-01}$ & $2.7\mathrm{e}{-02}$ \\
		$\Delta \gamma^{\text{p}}$ & - & $2.7\mathrm{e}{-04}$ & $9.1\mathrm{e}{-12}$ & $1.8\mathrm{e}{-03}$ & $3.0\mathrm{e}{-04}$ \\
		$p$ & kPa & $1.1\mathrm{e}{+04}$ & $1.0$ & $9.9\mathrm{e}{+04}$ & $1.5\mathrm{e}{+04}$ \\
		$\Delta p$ & kPa & $2.2\mathrm{e}{+02}$ & $-8.2\mathrm{e}{+01}$ & $2.8\mathrm{e}{+03}$ & $3.3\mathrm{e}{+02}$ \\
		$q$ & kPa & $2.5\mathrm{e}{+04}$ & $2.3$ & $2.3\mathrm{e}{+05}$ & $3.5\mathrm{e}{+04}$ \\
		$\Delta q$ & kPa & $4.9\mathrm{e}{+02}$ & $0.0$ & $6.0\mathrm{e}{+03}$ & $7.4\mathrm{e}{+02}$ \\
		$e$ & - & $5.6\mathrm{e}{-01}$ & $3.2\mathrm{e}{-01}$ & $7.6\mathrm{e}{-01}$ & $8.6\mathrm{e}{-02}$ \\
		$\Delta e$ & - & $-7.8\mathrm{e}{-04}$ & $-4.6\mathrm{e}{-03}$ & $1.9\mathrm{e}{-03}$ & $8.8\mathrm{e}{-04}$ \\
		\hline
	\end{tabular}
	\label{Table:4}
	\renewcommand{\arraystretch}{1}
\end{table}

\subsection{Regular and Physics-informed Architectures for Sands}\label{sec:WG-ANNs}
\begin{figure}
	\begin{center}
		\includegraphics[width=0.98\linewidth]{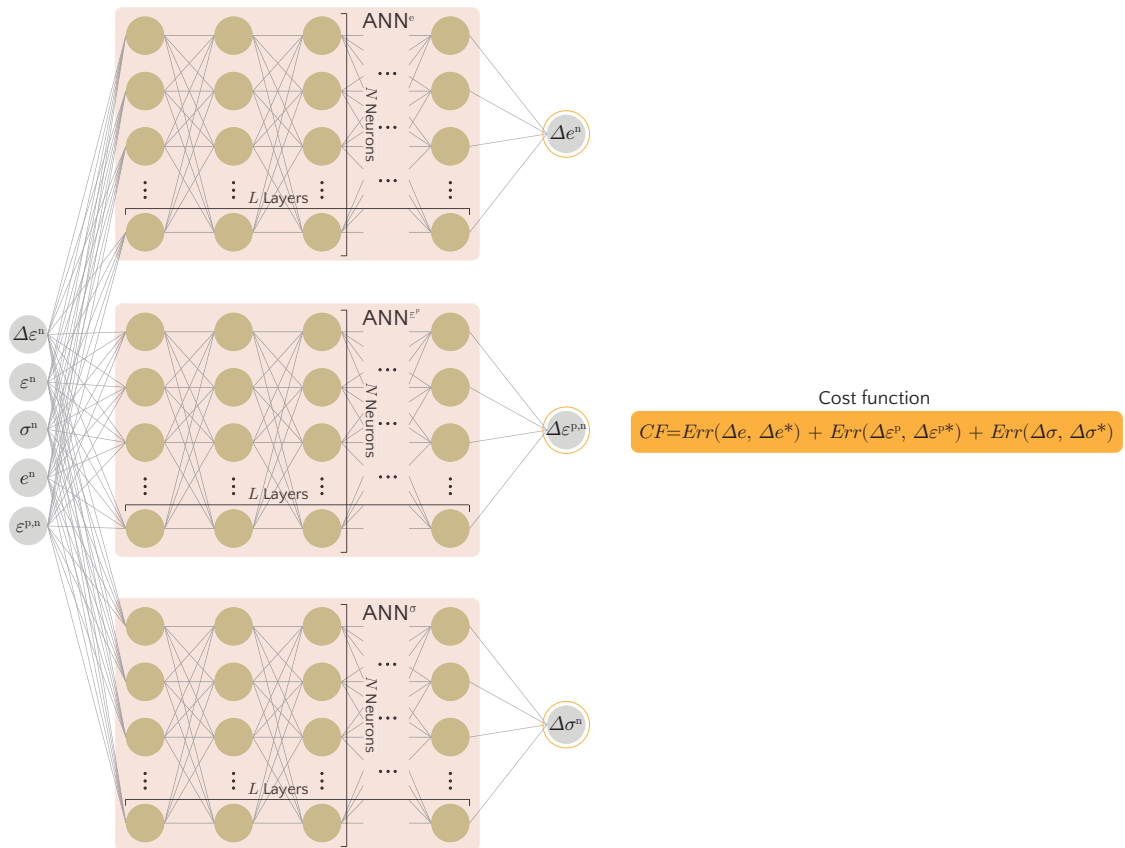}
		\caption{\label{fig:5} Parallel ANN architecture for modeling sands.}
	\end{center}
\end{figure}

\begin{figure}
	\begin{center}
		\includegraphics[width=0.98\linewidth]{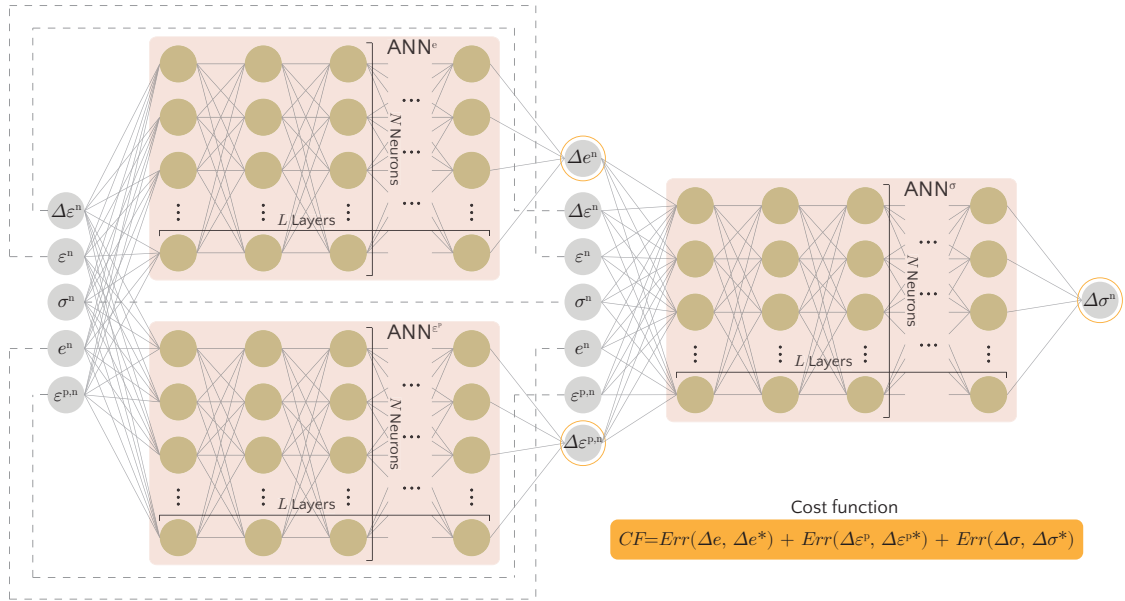}
		\caption{\label{fig:6} Serial ANN architecture for modeling sands.}
	\end{center}
\end{figure}

\begin{figure}
	\begin{center}
		\includegraphics[width=0.98\linewidth]{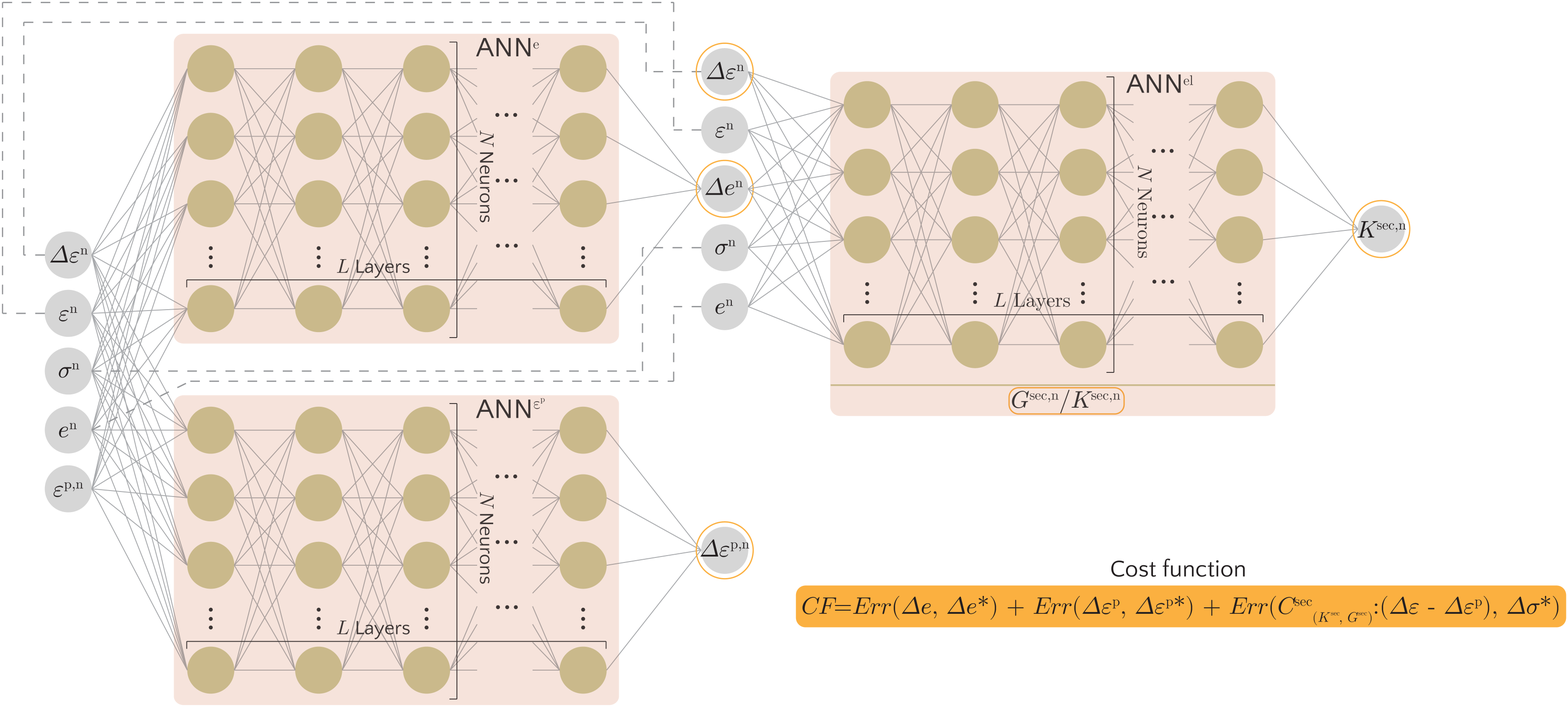}
		\caption{\label{fig:7} EPNN architecture for modeling sands.}
	\end{center}
\end{figure}

The neural network architectures presented in Section \ref{sec:ann-general} for a general elasto-plastic material are adapted here to sands. Figures \ref{fig:5} and \ref{fig:6} show the parallel and serial architectures for the regular ANN, while Fig. \ref{fig:7} shows the EPNN architecture. As can be seen in these figures, in all three architectures, we define two separate sub-networks for the two internal state variables, i.e. void ratio and plastic strain. Test trainings showed a better training can be achieved in this way compared to defining a single sub-network that predicts both of these internal state variables. This is inspired by and in agreement with the results presented by \citet{Lefik2003} where they defined separate sub-networks for prediction of each Cartesian component of stress. 

In the EPNN architecture, we enforced the isotropy of the material and replaced the full secant elasticity matrix with the secant bulk modulus $K^{\text{sec}}$ and secant shear modulus $G^{\text{sec}}$. Additionally, the ratio $G^{\text{sec}}/K^{\text{sec}}$ is taken to be constant (not dependent on the inputs of the network) and one of the parameters of the network to be optimized (alongside the link weights). This ratio is a function of secant Poisson's ratio ($\nu^{\text{sec}}$) which is usually considered to be constant for soils. Furthermore, the plastic strain and the predicted plastic strain increments in EPNN are not used as the inputs of the elasticity sub-network; since in uncoupled elastoplastic framework, the elasticity tensor is assumed to be independent of plastic strain.

\subsection{Choice of the Network Hyperparameters}\label{sec:network-params}
In this section, we present a systematic procedure for identifying the appropriate neural network hyperparameters. This important step is usually skipped in similar studies where the hyperparameters are chosen by trial and error.

Based on the discussion in Section \ref{sec:background-ann}, the parameters of neural network that need to be determined include: number of layers $L$, the number of neurons in each layer $N$, type of the activation functions, type of the loss function, and degree of regularization. Also, by comparing the predictions of the network on the training and cross-validation sets during optimization iterations (epochs), we should determine whether early-stopping is required or not.

The hyperparameters are identified for the parallel ANN architecture shown in Fig. \ref{fig:5}; the same parameters will then be used for the other two architectures when comparing the performance of different architectures. In the following, we start with using LeakyReLU activation function, no regularization, and no bias term in the output layer for all three sub-networks of the parallel ANN. In addition, we use the MAE loss function for training the network. By looking at the sub-networks' prediction error on the training and cross-validation sets, we first find the optimum number of layers and the number of neurons in each layer. Next, we look at the learning curve to decide whether regularization is needed for any of the sub-networks. After that, we check for the need to use early-stopping for any of the sub-networks. Finally, we study the effect of type of the activation function, and type of the loss function on the training behavior of the sub-networks.

For this study, we developed a neural network code based on the PyTorch library which is accessible online on the following repository:
\begin{equation*}
	\text{\href{https://github.com/meghbali/ANNElastoplasticity}{https://github.com/meghbali/ANNElastoplasticity}}
\end{equation*}

The repository includes the code for the EPNN architecture; the regular architectures can be constructed by straightforward modification of this code.

For all trainings in this section, the learning rates used in the ADAM optimizer are $0.0003$ for the stress sub-network, and $0.001$ for the plastic strain and void ratio sub-networks. Due to the presence of different variables with different range of values in both the input features and labels, all data (input features and labels) are normalized in the range $[-1.0, 1.0]$.

In order to truly measure the robustness of a network, the training data is usually divided into ``training set", ``cross-validation set" and ``test set". The training is done on the ``training set", while the ``cross-validation set" helps in tuning the network hyperparameters. Finally, the network is tested on the ``test set" to provide us with an idea of the generalization capability of the network. In this work, we randomly shuffle the data and use $60\%$ of it as the training set, $20\%$ as cross-validation set and the remaining $20\%$ as test set. The network is initialized by assigning random values to its weights \footnote{For reproducibility of the results and achieving consistency between different trainings, a fixed randomization seed is used for generating random numbers in the program.}. As the network is trained, its weights are adjusted with the goal of minimizing the cost function.

\subsubsection{Number of Layers and Neurons}\label{sec:layers-neurons}
In this work, we use the same number of neurons for each layer of neural networks. We start by finding the optimum number of layers assuming $50$ neurons in each layer. Other parameters of the networks are the same as the ones mentioned at the beginning of Section \ref{sec:network-params}.

\begin{figure}[t]
	\begin{center}
		\includegraphics[width=0.55\linewidth]{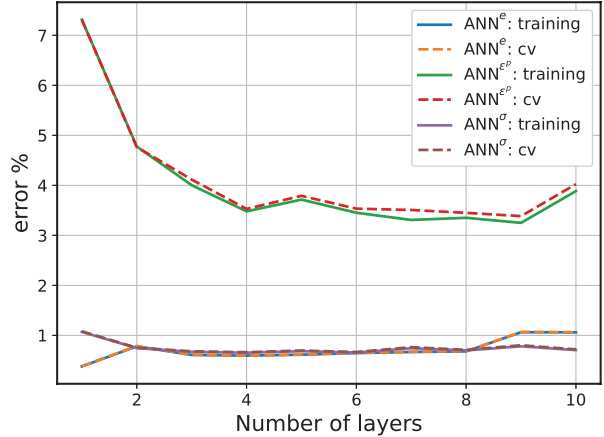}
		\caption{\label{fig:8} Prediction errors on the training and cross-validation (cv) sets after $1e4$ epochs of training versus number of layers, for the sub-networks of the parallel ANN architecture. The number of neurons in each layer is taken as $50$ for all three sub-networks.}
	\end{center}
\end{figure}

The error in this work is defined based on the error in the Frobenius norm of the output matrix of each sub-network with respect to the labeled data. The final error on the training and cross-validation sets after $1e4$ epochs of training versus the number of layers is shown in Fig. \ref{fig:8}. It can be seen in this figure that the prediction error on the cross-validation set is always very close to the prediction error on the test set for all three sub-networks. This shows the network's satisfactory generalization capability which, as will be shown in Section \ref{sec:learning-curve}, is tied to the size of the training set; as we use less data for training the network, the network gradually loses its generalization capability, resulting in higher prediction error on the cross-validation set.
For both the void ratio and stress sub-networks, the errors are mostly below $1\%$ for all number of layers. We choose $3$ layers for both sub-networks, as increasing the number of layers beyond $3$ has no significant effect on their training. On the other hand, a more significant final error is observed in Fig. \ref{fig:8} for the plastic strain sub-network. Based on this figure, we use $4$ layers for the plastic strain sub-network, as using more layers offers only a marginal improvement on the training error of the plastic strain sub-network.
 
\begin{figure}[t]
	\begin{center}
		\includegraphics[width=0.55\linewidth]{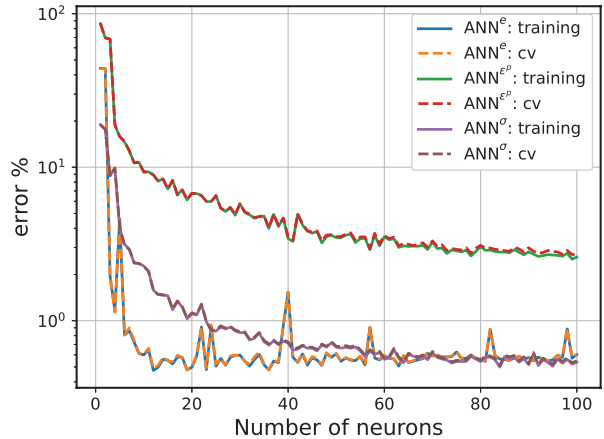}
		\caption{\label{fig:10} Prediction errors on the training and cross-validation (cv) sets after $1e4$ epochs of training versus the number of neurons in each layer, for the sub-networks of the parallel ANN architecture. The number of layers is equal to $3$ for the void ratio and stress sub-networks, and $4$ for the plastic strain sub-network.}
	\end{center}
\end{figure}

Using $3$ layers for the void ratio and stress sub-networks, and $4$ layers for the plastic strain sub-network, we now plot the final error on the training and cross-validation sets after $1e4$ epochs of training versus the number of neurons in each layers in Fig. \ref{fig:10}. For both the void ratio and stress sub-networks, the errors are below $1\%$ for more than $25$ neurons and stabilize at $60$ neurons. Therefore, for both networks we choose $60$ neurons in each layer. For the plastic strain sub-network, the improvement beyond $75$ neurons is marginal and as such, we use $75$ neurons for the plastic strain sub-network.

\begin{figure}
	\begin{center}
		\includegraphics[width=0.55\linewidth]{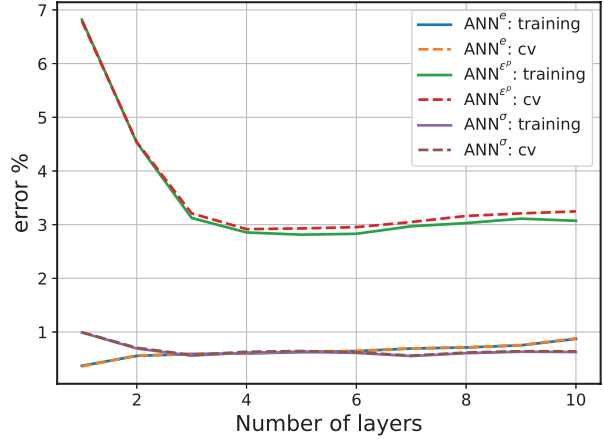}
		\caption{\label{fig:11} Prediction errors on the training and cross-validation (cv) sets after $1e4$ epochs of training versus number of layers, for the sub-networks of the parallel ANN architecture. The number of neurons in each layer is taken as $60$ for the void ratio and stress sub-networks, and $75$ for the plastic strain sub-network.}
	\end{center}
\end{figure}

In order to cross-check the choice of layers and neurons for the sub-networks, we plot the errors-vs-number of layers for the sub-networks, this time considering $60$ neurons in each layer for the void ratio and stress sub-networks, and $75$ neurons in each layer for the plastic strain sub-network (see Fig. \ref{fig:11}). Looking at the errors on the training and cross-validation sets, the choice of $3$ layers for the void ratio and stress sub-networks, and $4$ layers for the plastic strain sub-network is confirmed.

\subsubsection{Learning Curve}\label{sec:learning-curve}
\begin{figure}
	\begin{center}
		\includegraphics[width=0.55\linewidth]{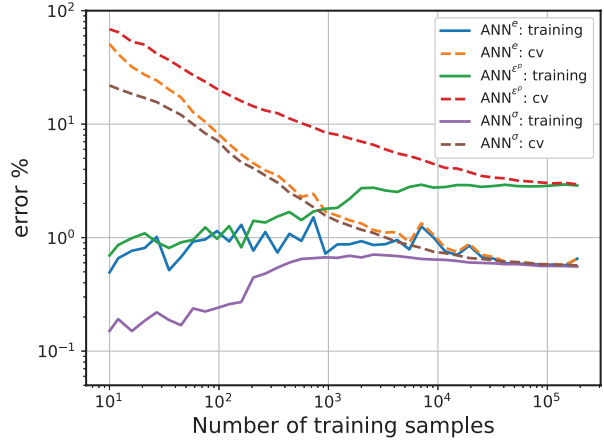}
		\caption{\label{fig:12} Learning curves for the void ratio, plastic strain and stress sub-networks in the parallel ANN architecture (based on $1e4$ epochs of training).}
	\end{center}
\end{figure}

The learning curve of a neural network demonstrates its learning process through the evolution of its generalization capability as more training data is presented to it. Figure \ref{fig:12} shows such curve for the parallel ANN architecture. The number of layers and neurons for each sub-network is based on the optimized values selected in Section \ref{sec:layers-neurons}. Other parameters of the sub-networks are the same as the ones mentioned at the beginning of Section \ref{sec:network-params}. The figure illustrates the prediction error on training and cross-validation sets after $1e4$ epochs of training versus the number of training samples (training set size). The prediction on the training set is based on the same data used for training, while the prediction on the cross-validation is based on the entire cross-validation set (a fixed $20\%$ of total data). For each training set size, $5$ different training sets are randomly selected from the complete training set and errors are calculated based on the average over the $5$ trainings.
In Fig. \ref{fig:12}, the learning curves are shown for all three sub-networks. It is seen that at small number of training samples, for all three sub-networks the error on the training set is small, while the error on the cross-validation set is large. This follows the intuition, as it is easy for the network to learn a small training set; however, there is not enough information in the training data to give acceptable generalization capability to the network. As more training data is provided to the network, it becomes increasingly harder for the network to learn the training data; but it can generalize better. That is the reason the prediction errors on the training set increase, while the ones on the cross-validation sets decrease. It is seen that the two converge for all three sub-networks. 

For all three sub-networks, the final error (corresponding to the use of full training data) is acceptable which shows the sub-networks are well-designed. None of the sub-networks suffer from ``high bias" or ``high variance" problems; thus, no regularization is needed, and the number of training samples is certainly enough. In fact, the learning curves of the void ratio and stress sub-networks show that as low as $25\%$ of the data is enough for the training of these two sub-networks. To further decrease the converged error of the plastic strain sub-network, additional features can be considered in the input data. However, this process involves trial and error in including different features which is beyond the scope of the current work.

\subsubsection{Training Curve}
\begin{figure}
	\begin{center}
		\includegraphics[width=0.55\linewidth]{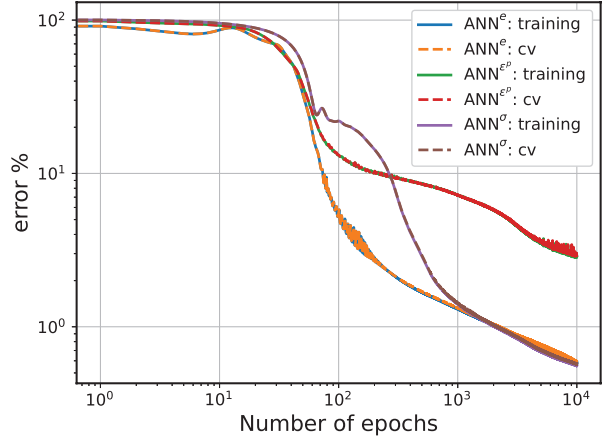}
		\caption{\label{fig:13} Training curve for the void ratio, plastic strain and stress sub-networks in the parallel ANN architecture.}
	\end{center}
\end{figure}

Figure \ref{fig:13} shows the training curves for sub-networks of the parallel ANN architecture with number of layers and neurons obtained in Section \ref{sec:layers-neurons}, and for the complete set of training data ($60\%$ of total data). It is seen that the training and cross-validation errors are very close during the optimization iterations. Hence, no early-stopping is required for any of the sub-networks.

\subsubsection{Type of Activation Function}
\begin{figure}
	\begin{center}
		\includegraphics[width=1.0\linewidth]{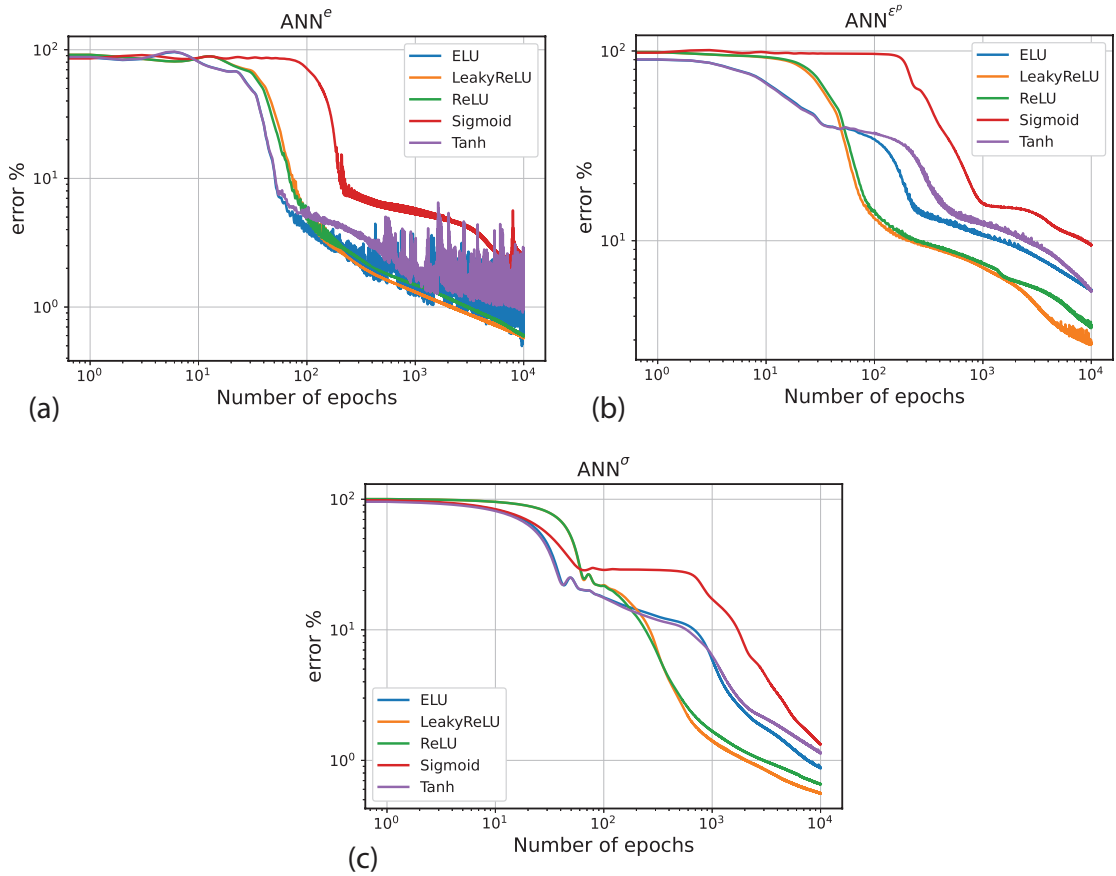}
		\caption{\label{fig:9} Effect of type of activation function on the training of sub-networks in the parallel ANN architecture.}
	\end{center}
\end{figure}

The effect of different activation functions, including ReLU, LeakyReLU, Sigmoid, Tanh, and ELU, on the training of the parallel ANN architecture is studied here. Details of these activation functions can be found in \cite{Masi2021b}, among others. Figure \ref{fig:9} clearly shows that the LeakyReLU activation function provides faster training for all three sub-networks over the other activation functions.

\subsubsection{Type of Loss Function}
\begin{figure}
	\begin{center}
		\includegraphics[width=1.0\linewidth]{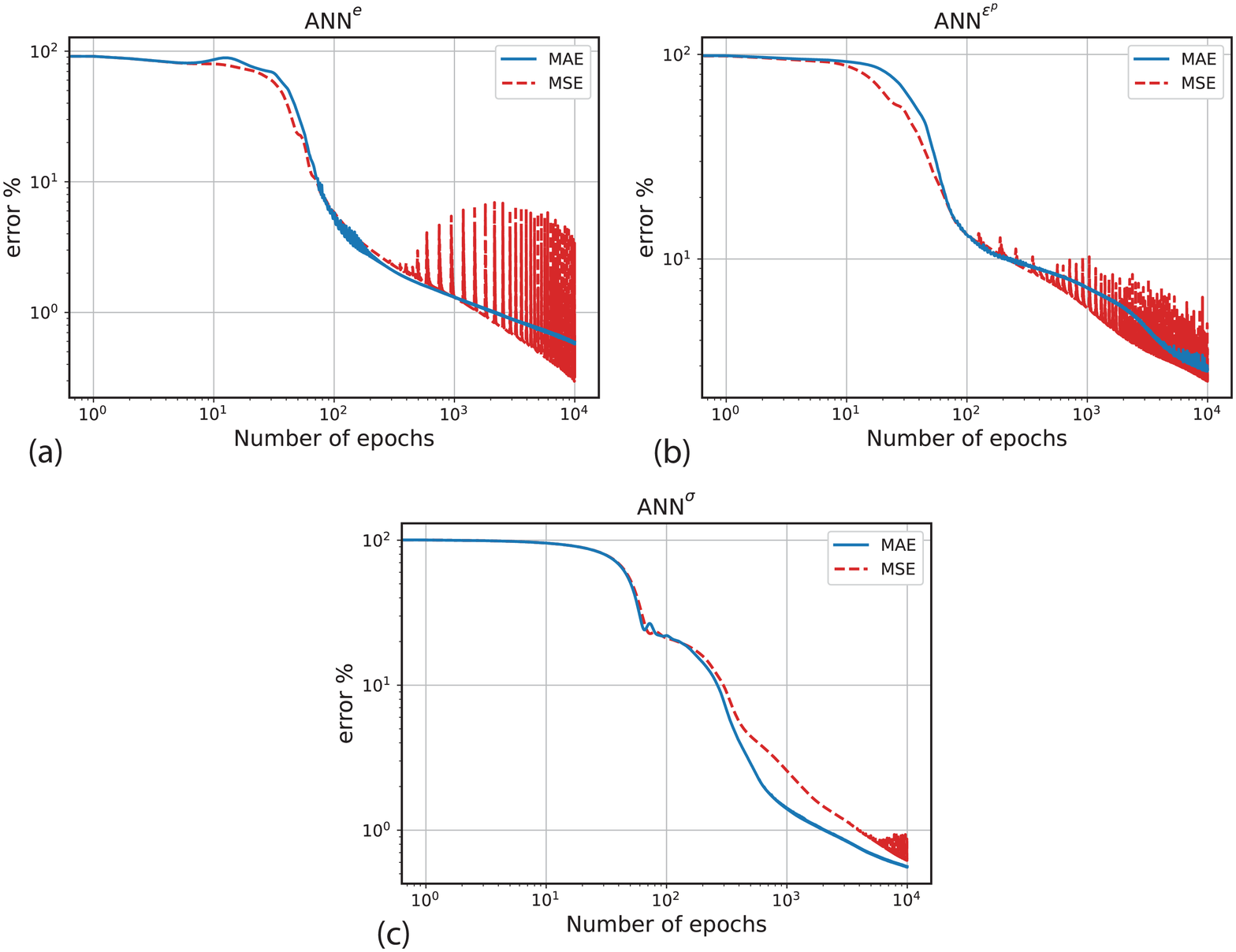}
		\caption{\label{fig:14} Effect of type of loss function on the training of sub-networks in the parallel ANN architecture.}
	\end{center}
\end{figure}

Figure \ref{fig:14} shows how training of the parallel ANN architecture is affected by the error (loss) function, i.e. MSE or MAE. The results indicate that the MAE loss function provides more stable training with less fluctuations for all three sub-networks. Moreover, using MSE loss function introduces an undesirable bias towards outputs with larger magnitudes, as smaller values will have less effect on the cost function and thus training of the network. In contrast, MAE loss function provides a more homogeneous weight to network outputs with varying magnitudes which is particularly important for predictions on the network in the stress states close to zero.

\subsection{Learning and Training Curves for Different Network Architectures}
In this section, we present the training results of the three neural network architectures presented in Section \ref{sec:WG-ANNs} for learning the elasto-plastic behavior of sands. In order to provide a systematic comparison, we use the same hyperparameters obtained in Section \ref{sec:network-params} for all the three architectures. The learning rate of the sub-networks are also kept the same; except, for the elasticity (stress) sub-network of EPNN, we use the learning rate of $0.01$ for sub-network weights and $0.001$ for the shear to bulk moduli ratio. The same dataset is used for training of all three network architectures which is pre-processed following the same procedure outlined at the beginning of Section \ref{sec:network-params}.

In order to give readers an idea of how much time-consuming the training of these networks are, the runtimes for $1e4$ epochs of training of EPNN is presented in Table \ref{Table:2} for sample processing units. Six gigabytes of RAM is sufficient to handle the computations for $190,000$ training samples. As mentioned before, the code developed for this study is based on the PyTorch library which can be run on GPU. As Table \ref{Table:2} suggests, the high performance GPU used in this study for model training provides roughly $100$ times faster training speeds over high-end CPUs in portable computers. 
\begin{table}[h]
	\scriptsize
	\renewcommand{\arraystretch}{1.2}
	\caption{Training runtimes of EPNN on $190,000$ training samples for selected processing units.}
	\centering
	\begin{tabular}{c c}
		\hline
		Processing unit & Runtime (seconds/$1,000$ epochs) \\ \hline
		Intel(R) Core(TM) i7-7500U CPU @ 2.70GHz & 5020 \\
		(Kaby Lake, 2017 microarchitecture) with 16 GB RAM & \\
		Intel(R) Xeon(R) Gold 6148 CPU @ 2.40GHz & 4730 \\
		(Skylake, 2019 microarchitecture) with 754 GB RAM & \\
		NVIDIA Tesla V100-PCIE GPU with 16 GB RAM & 55 \\
		\hline
	\end{tabular}
	\label{Table:2}
	\renewcommand{\arraystretch}{1}
\end{table}

\begin{figure}[t]
	\begin{center}
		\includegraphics[width=1.0\linewidth]{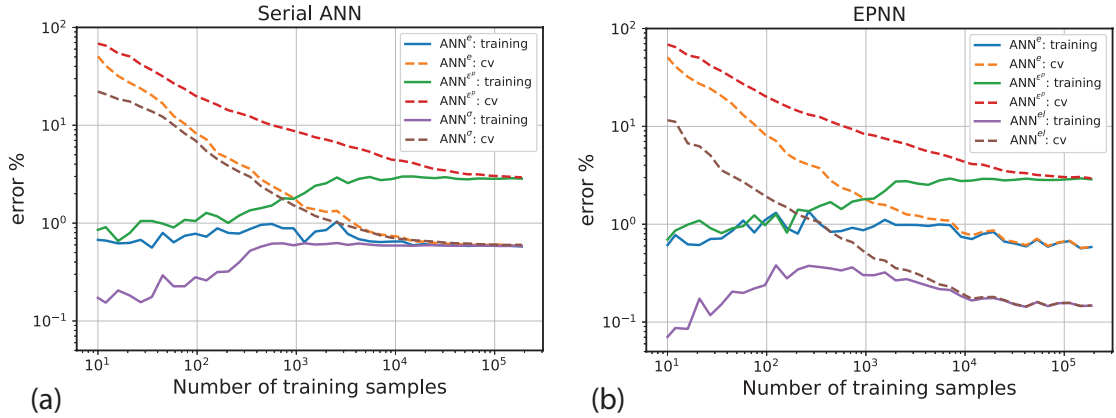}
		\caption{\label{fig:15} Learning curves for the (a) serial ANN and (b) EPNN architectures.}
	\end{center}
\end{figure}

\begin{figure}
	\begin{center}
		\includegraphics[width=0.55\linewidth]{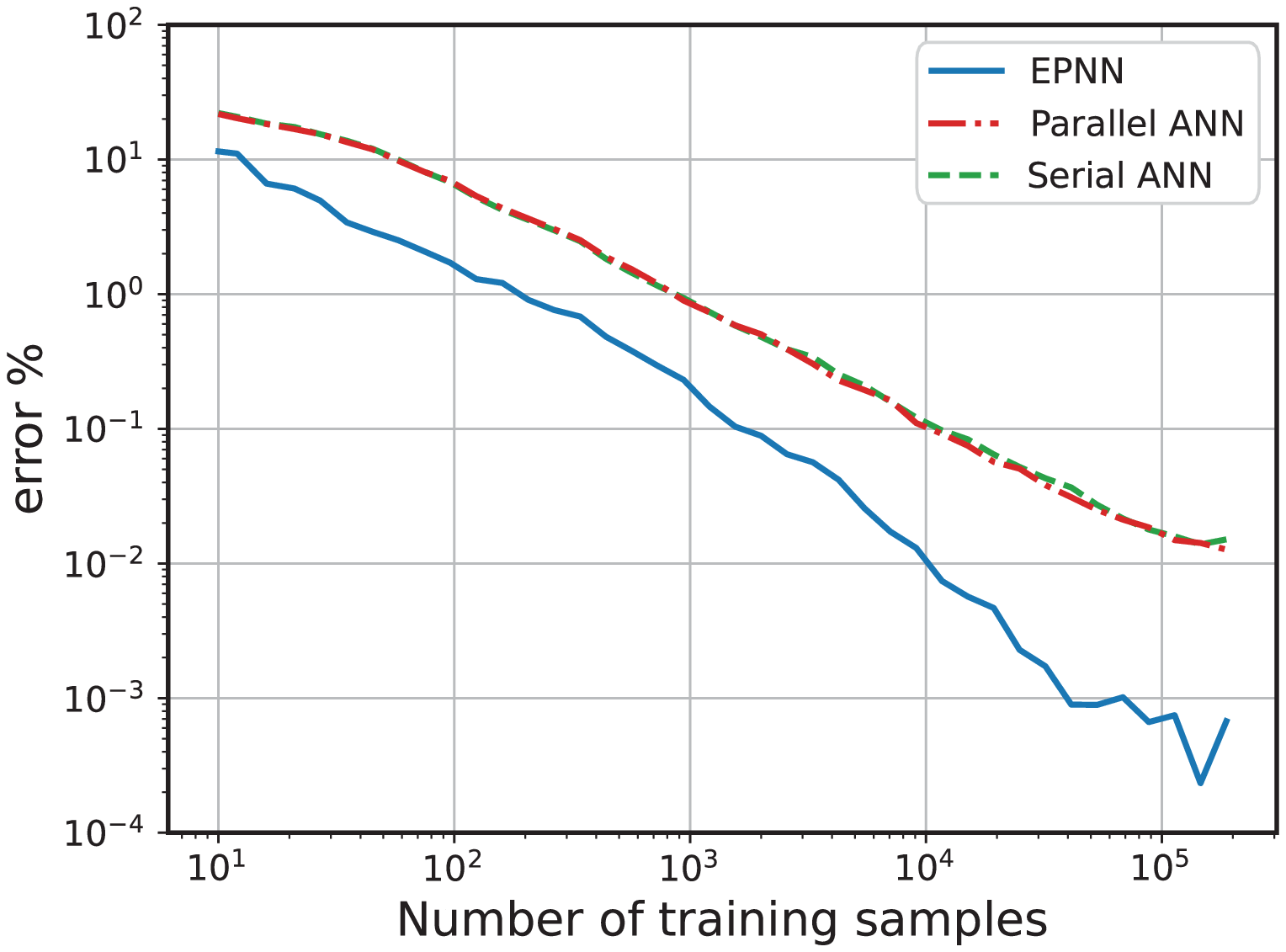}
		\caption{\label{fig:33} Difference between the training and cross-validation errors versus the number of training samples, for the three network architectures considered for learning elasto-plasticity.}
	\end{center}
\end{figure}

The learning curves of the serial ANN and EPNN are presented in Fig. \ref{fig:15}, while the ones for the parallel ANN architecture were given earlier in Fig. \ref{fig:12}. It is seen from the comparison of the learning curves that the parallel and serial architectures train in the same way with their final errors being virtually the same. In the EPNN architecture, while the training of the void ratio and plastic strain is similar to the other two architectures, the training of the stress is significantly improved. Figure \ref{fig:33} presents a closer look at the difference between the training and cross-validation errors for the stress sub-network in the three architectures. Based on Figs. \ref{fig:15} and \ref{fig:33}, the difference between the training and cross-validation errors in the stress sub-network in EPNN converges faster to a lower final error. If a tolerance of $0.01$ is chosen for convergence in Fig. \ref{fig:33}, EPNN achieves the same training accuracy for the stress with at least one order of magnitude less training data. Hence, if additional physics are added to EPNN in relation to the prediction of plastic strain, EPNN can potentially be trained with a fraction of data and provide improved predictive capability on the unseen data over the two ANN architectures. This could involve physics pertaining specifically to the plastic potential and yield function.

\begin{figure}
	\begin{center}
		\includegraphics[width=1.0\linewidth]{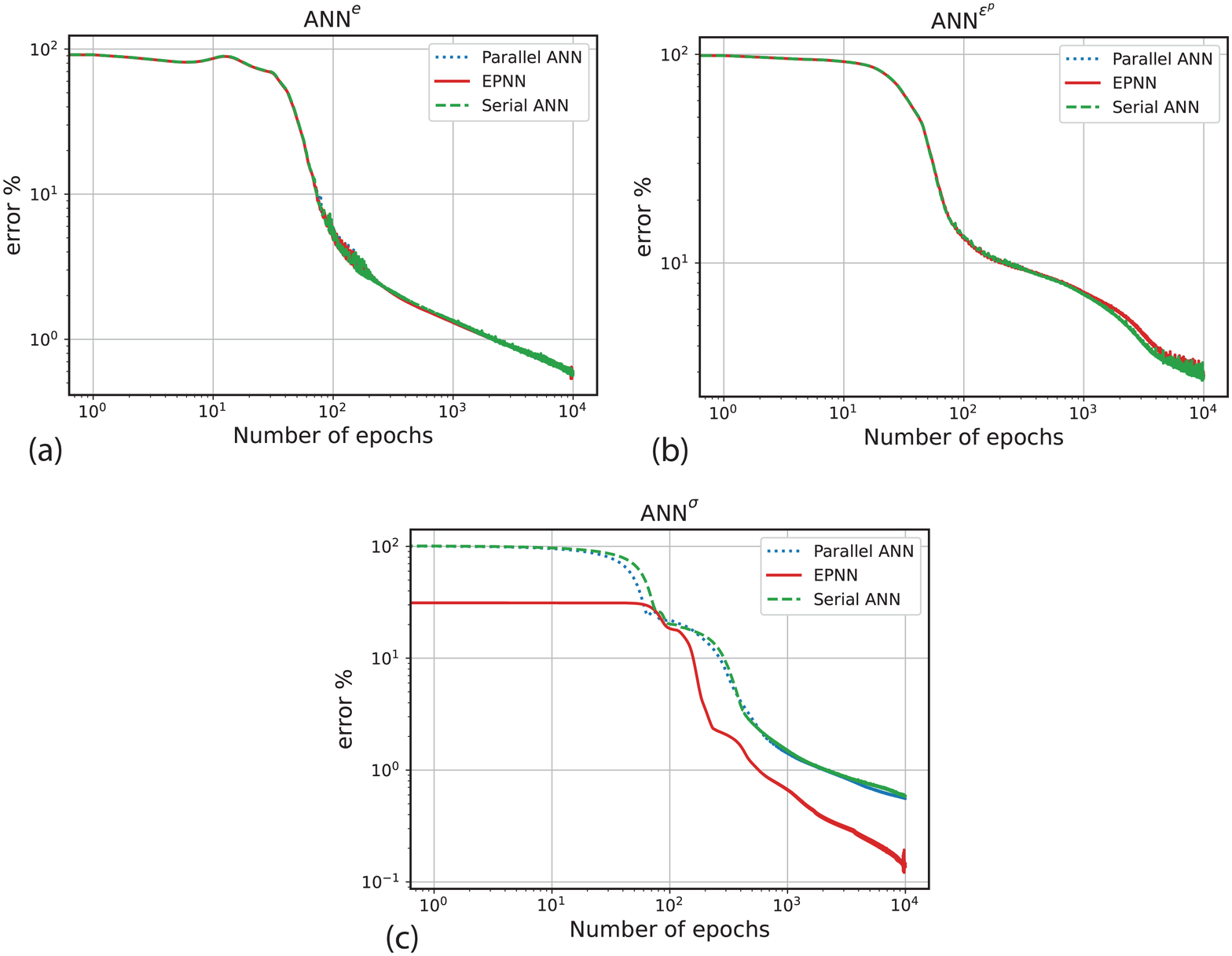}
		\caption{\label{fig:16} Comparison of the training curves of different network architectures for the (a) void ratio, (b) plastic strain, and (c) stress sub-networks.}
	\end{center}
\end{figure}

Figure \ref{fig:16} presents the comparison between the training of different sub-networks for the parallel ANN, serial ANN and EPNN architectures. It is seen in this figure that the training of the void ratio and plastic strain are virtually the same in all three architectures, whereas training of the stress sub-network in EPNN is improved due to its physics-informed architecture. This means EPNN can learn the training data better. It should be pointed out that while the final errors of the two regular ANN architectures for the stress network are also acceptable, the results in Figure \ref{fig:16} only demonstrate the achieved accuracy of the models in learning the isolated training samples; their ability to predict complete loading paths and unseen cases will be explored in the next section.

\section{Application to Plasticity of Sands: Verification}\label{sec:verification}
In this section, the predictive capabilities of the three network architectures for elasto-plasticity introduced in Section \ref{sec:WG-ANNs} are compared in several examples. 

\subsection{Predictions on Test Set}
First, the prediction errors of the sub-networks of each architecture on the unseen test set are compared in Table \ref{Table:3}. It is seen that all three architectures provide a very good prediction on the test set, with EPNN offering a better prediction for the stresses.
\begin{table}[h]
	\scriptsize
	\renewcommand{\arraystretch}{1.2}
	\caption{Prediction errors on the test set.}
	\centering
	\begin{tabular}{c | c c c}
		\hline
		\backslashbox[12mm]{sub-network}{architecture} & Parallel ANN & Serial ANN & EPNN \\ \hline
		ANN$^{\text{e}}$ & $0.59\%$ & $0.59\%$ & $0.59\%$ \\
		ANN$^{\upvarepsilon^{\text{p}}}$ & $2.98\%$ & $2.91\%$ & $2.98\%$ \\
		ANN$^{\upsigma}$, ANN$^{\text{el}}$ & $0.58\%$ & $0.59\%$ & $0.14\%$ \\
		\hline
	\end{tabular}
	\label{Table:3}
	\renewcommand{\arraystretch}{1}
\end{table}

\subsection{Predictions on a Complete Loading Path in Training Data}
Next, we use the trained networks in recall mode to simulate complete strain-controlled loading paths. The verification here is particularly different from those in Table \ref{Table:3} in that here, the trained network is called multiple times and the outputs of each step is used in updating stresses, strains, and state variables, which are then used as feedback input to the network for the following step.
\begin{algorithm}[t]
	\caption{Using the trained networks in recall mode for strain-controlled material point tests.}\label{alg:1}
	\begin{algorithmic}[1]
		\Require Pre-trained networks. number of loading steps (nstep).\\
		$\boldsymbol{\varepsilon}^0, \boldsymbol{\sigma}^0, e^0, \boldsymbol{\varepsilon}^{\text{p},0}$ $\gets$ Initialize state variables.\\
		$\text{n}$ $\gets$ $0$
		\For{istep = 1, nstep}\\
		$\Delta \boldsymbol{\varepsilon}^{\text{n}}$ $\gets$ Prescribe strain increment.\\
		$\Delta e^{\text{n}}, \Delta \boldsymbol{\varepsilon}^{\text{p},\text{n}}, \Delta \boldsymbol{\sigma}^{\text{n}}$ $\gets$ Calculate output of the Neural Network, given the inputs to the network: $\boldsymbol{\varepsilon}^{\text{n}}, \boldsymbol{\sigma}^{\text{n}},e^{\text{n}},\boldsymbol{\varepsilon}^{\text{p},\text{n}}$ and $\Delta \boldsymbol{\varepsilon}^{\text{n}}$.\\
		$\boldsymbol{\varepsilon}^{\text{n}+1}$ $\gets$ $\boldsymbol{\varepsilon}^{\text{n}}+\Delta \boldsymbol{\varepsilon}^{\text{n}}$, $\boldsymbol{\sigma}^{\text{n}+1}$ $\gets$ $\boldsymbol{\sigma}^{\text{n}}+\Delta \boldsymbol{\sigma}^{\text{n}}$, $e^{\text{n}+1}$ $\gets$ $e^{\text{n}}+\Delta e^{\text{n}}$, $\boldsymbol{\varepsilon}^{\text{p},\text{n}+1}$ $\gets$ $\boldsymbol{\varepsilon}^{\text{p},\text{n}}+\Delta \boldsymbol{\varepsilon}^{\text{p},\text{n}}$ Update the state variables.\\
		$\text{n}$ $\gets$ $\text{n}+1$
		\EndFor
	\end{algorithmic}
\end{algorithm}
Such a feedback test is crucial in evaluating constitutive models, since it explores the possibility of progressively increasing error that can eventually deviate the response away from the ground truth. This provides us with a better understanding of the robustness and extrapolation capabilities of different architectures and how they compare with each other. The algorithm for the use of neural networks in recall mode is presented in Algorithm \ref{alg:1}. As the algorithm suggests, one loading step using the neural network is significantly simpler to implement and less computationally expensive than the classical elasto-plasticity which requires implementing a return mapping algorithm \citep{Simo1998}.

\begin{figure}
	\begin{center}
		\includegraphics[width=1.0\linewidth]{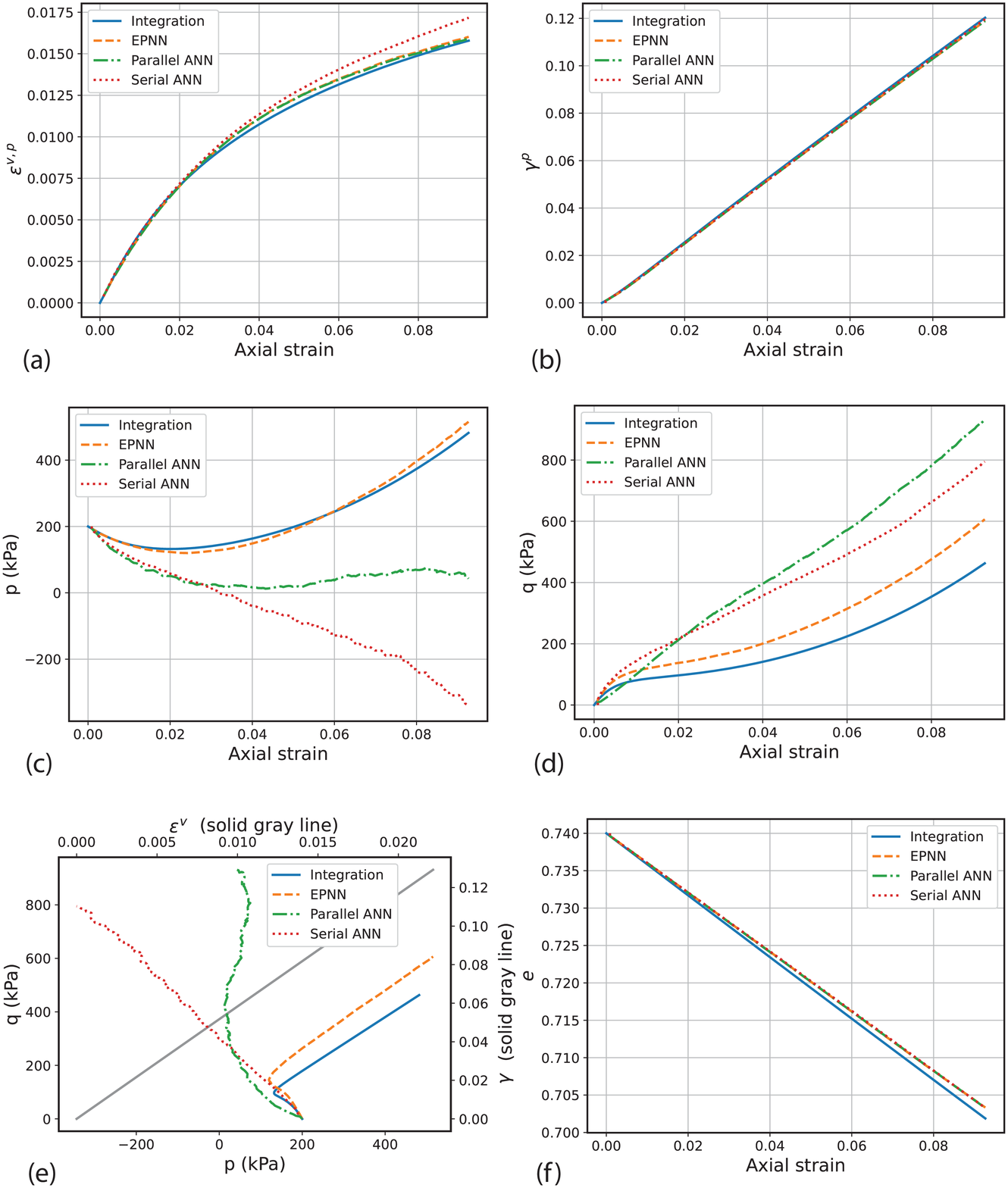}
		\caption{\label{fig:24} Comparison between the predictions of the three network architectures and the ground truth (numerical integration results) for (a) volumetric plastic strain, (b) equivalent plastic shear strain, (c) mean stress, (d) equivalent shear stress, (e) strain (solid gray line) and stress paths, and (f) void ratio, in one of the loading paths used in the training dataset with $p^{\text{in}}=200$ kPa and $e^{\text{in}}=0.74$.}
	\end{center}
\end{figure}

For the first test, we choose one of the loading paths used in training the networks with $p^{\text{in}}=200$ kPa and $e^{\text{in}}=0.74$. The strain path followed in this verification is shown in Fig. \ref{fig:24}e (solid gray line). Figure \ref{fig:24} shows the comparison between the predictions of the three architectures and the ground truth. The ``Axial Strain" in these figures refers to $\varepsilon_{33}$ component of the strain tensor. The results in Table \ref{Table:3} and Fig. \ref{fig:24} indicate that although all the three architectures provide very good predictions over the entire dataset, their predictions on a complete loading path included in the dataset is far from being perfect. The performance of the EPNN, on the other hand, is remarkably superior to both regular ANNs. Regardless of the model being recalled with feedback features multiple times, the response retains the characteristics of the ground truth and the errors do not accumulate. While part of this superiority can be attributed to the additional accuracy of the EPNN (shown in Table \ref{Table:3}), the more accurate prediction of the EPNN in Fig. \ref{fig:24} is due to its enhanced generality which is due directly to the physics hardwired into the architecture of EPNN \footnote{For a detailed discussion on the difference between accuracy and generality, the readers are referred to \citet{Pouragha2020}}.

\subsection{Predictions on the Undrained Loading Path}
\begin{figure}
	\begin{center}
		\includegraphics[width=0.55\linewidth]{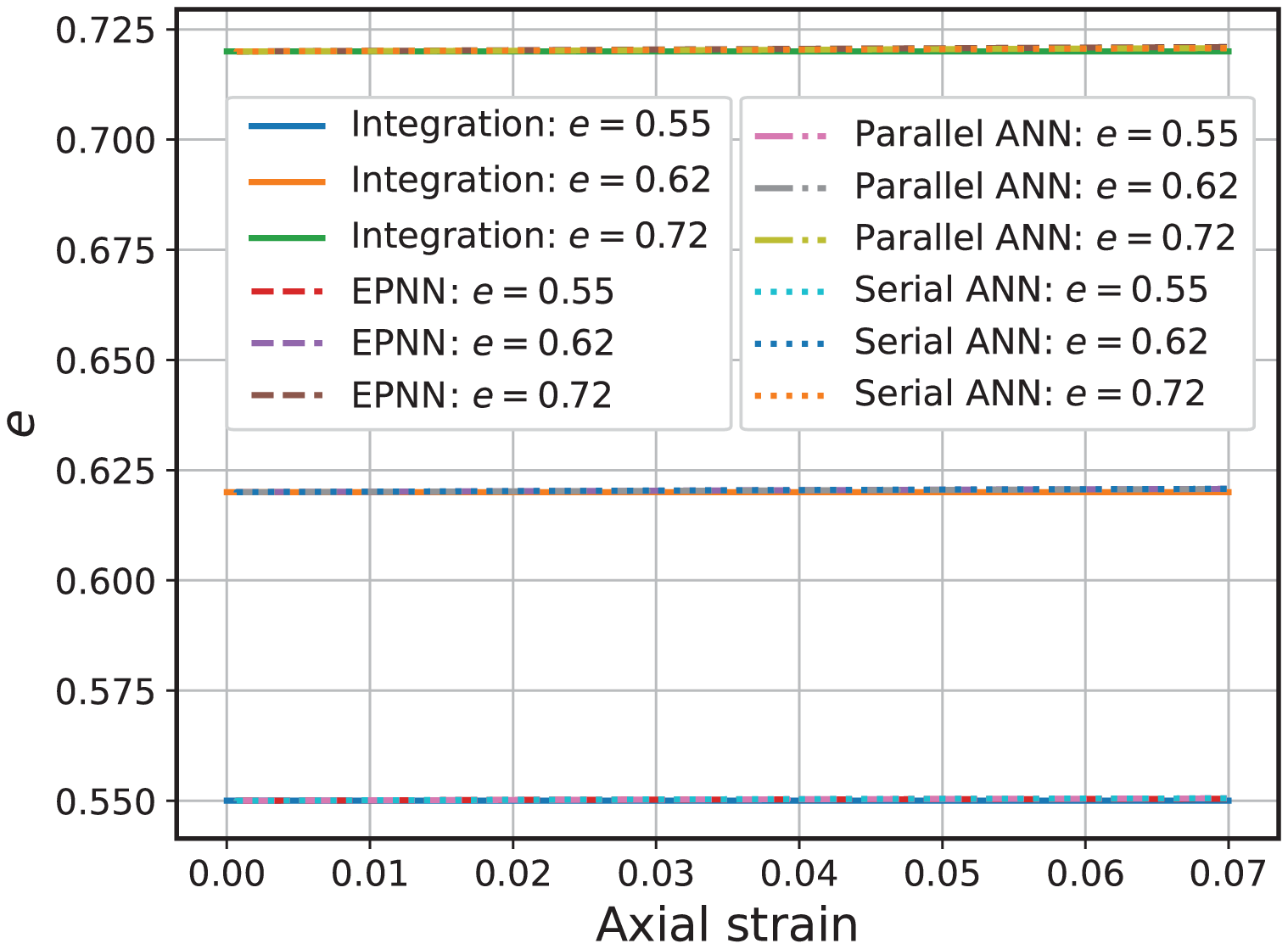}
		\caption{\label{fig:21} Comparison between the predictions of the three network architectures and the ground truth (numerical integration results) for void ratio, in the undrained test with $p^{\text{in}}=225$ kPa and different initial void ratios.}
	\end{center}
\end{figure}

\begin{figure}
	\begin{center}
		\includegraphics[width=1.0\linewidth]{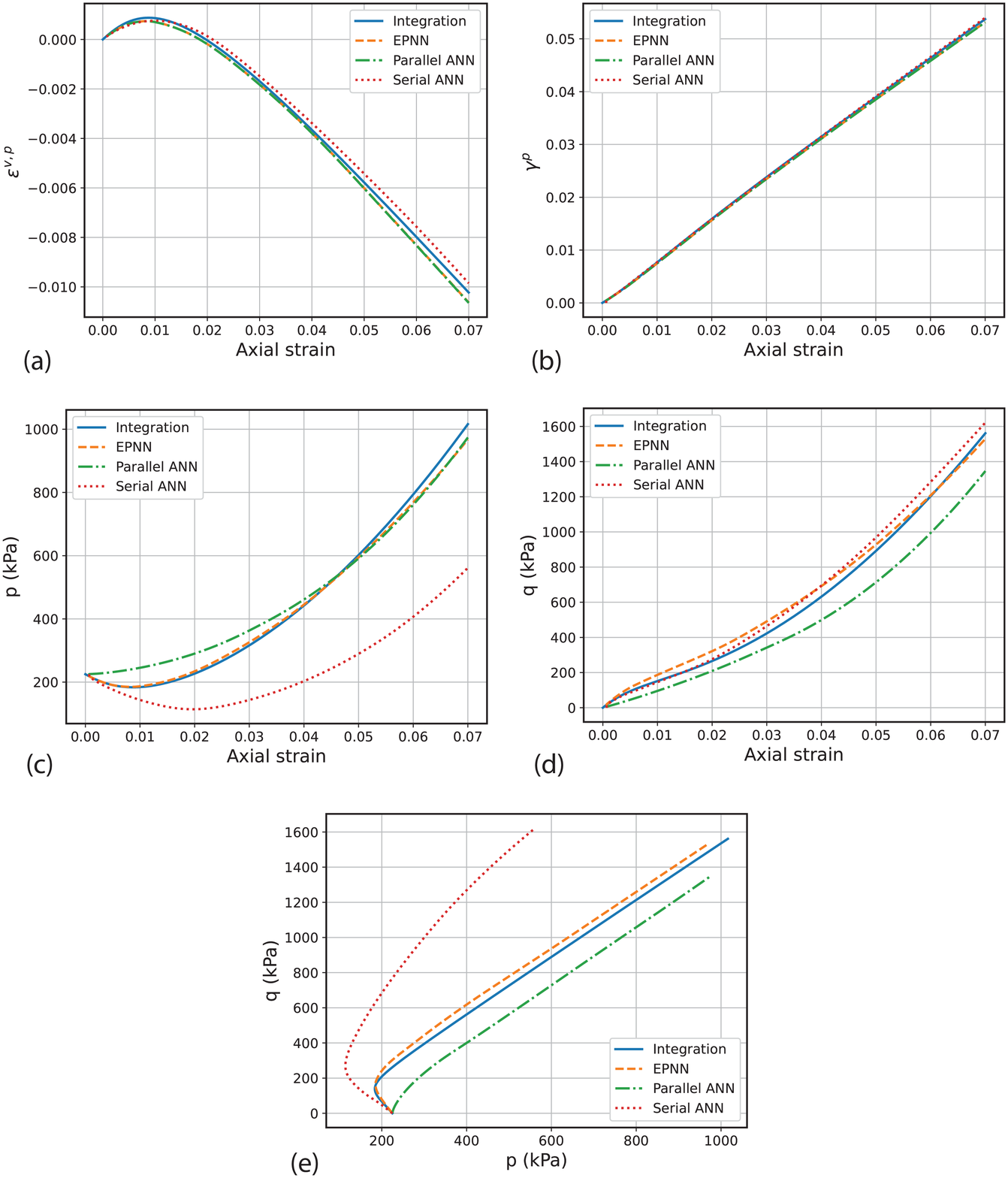}
		\caption{\label{fig:18} Comparison between the predictions of the three network architectures and the ground truth (numerical integration results) for (a) volumetric plastic strain, (b) equivalent plastic shear strain, (c) mean stress, (d) equivalent shear stress, and (e) stress path, in the undrained test with $p^{\text{in}}=225$ kPa and $e^{\text{in}}=0.55$.}
	\end{center}
\end{figure}

\begin{figure}
	\begin{center}
		\includegraphics[width=1.0\linewidth]{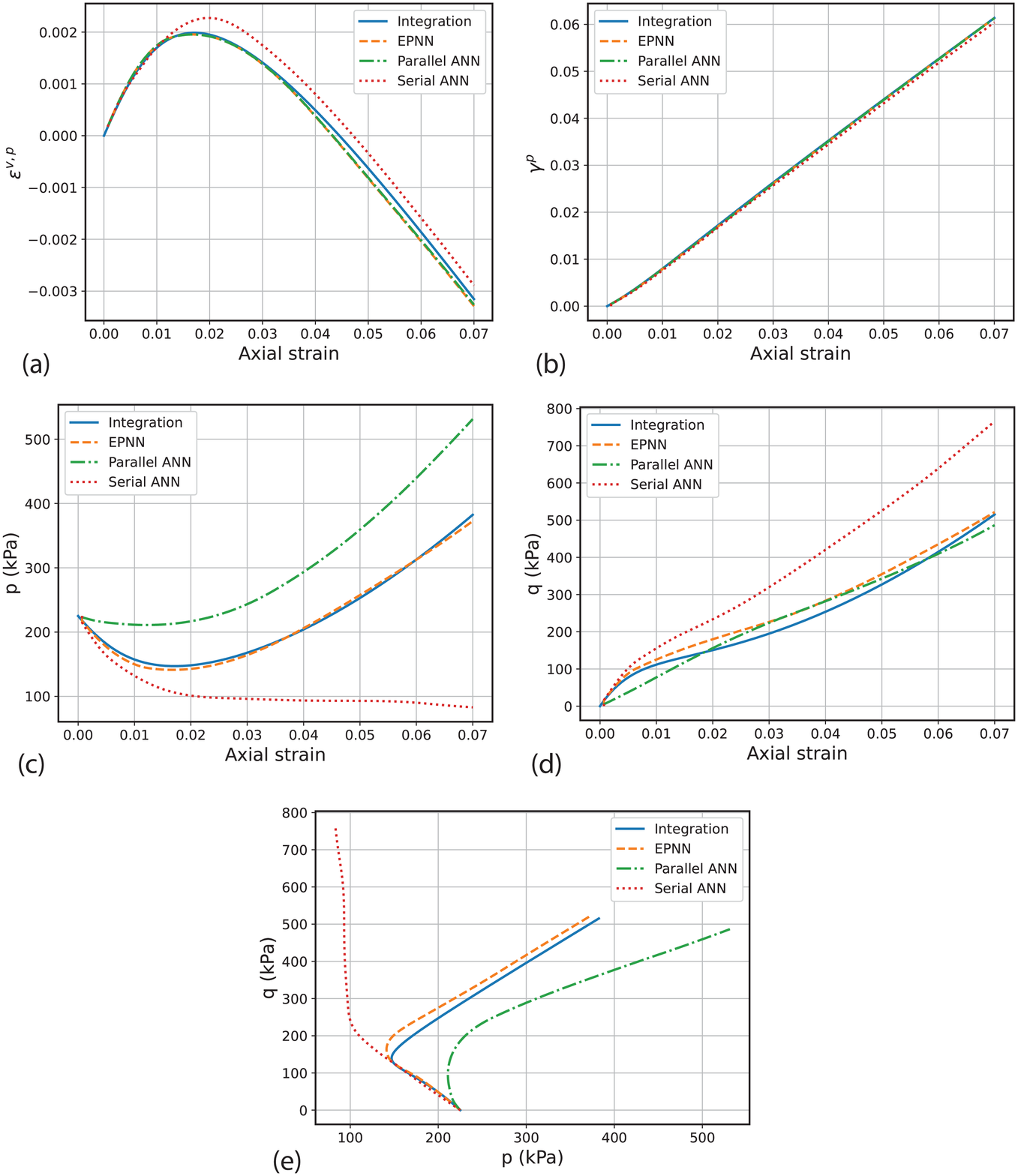}
		\caption{\label{fig:19} Comparison between the predictions of the three network architectures and the ground truth (numerical integration results) for (a) volumetric plastic strain, (b) equivalent plastic shear strain, (c) mean stress, (d) equivalent shear stress, and (e) stress path, in the undrained test with $p^{\text{in}}=225$ kPa and $e^{\text{in}}=0.62$.}
	\end{center}
\end{figure}

\begin{figure}
	\begin{center}
		\includegraphics[width=1.0\linewidth]{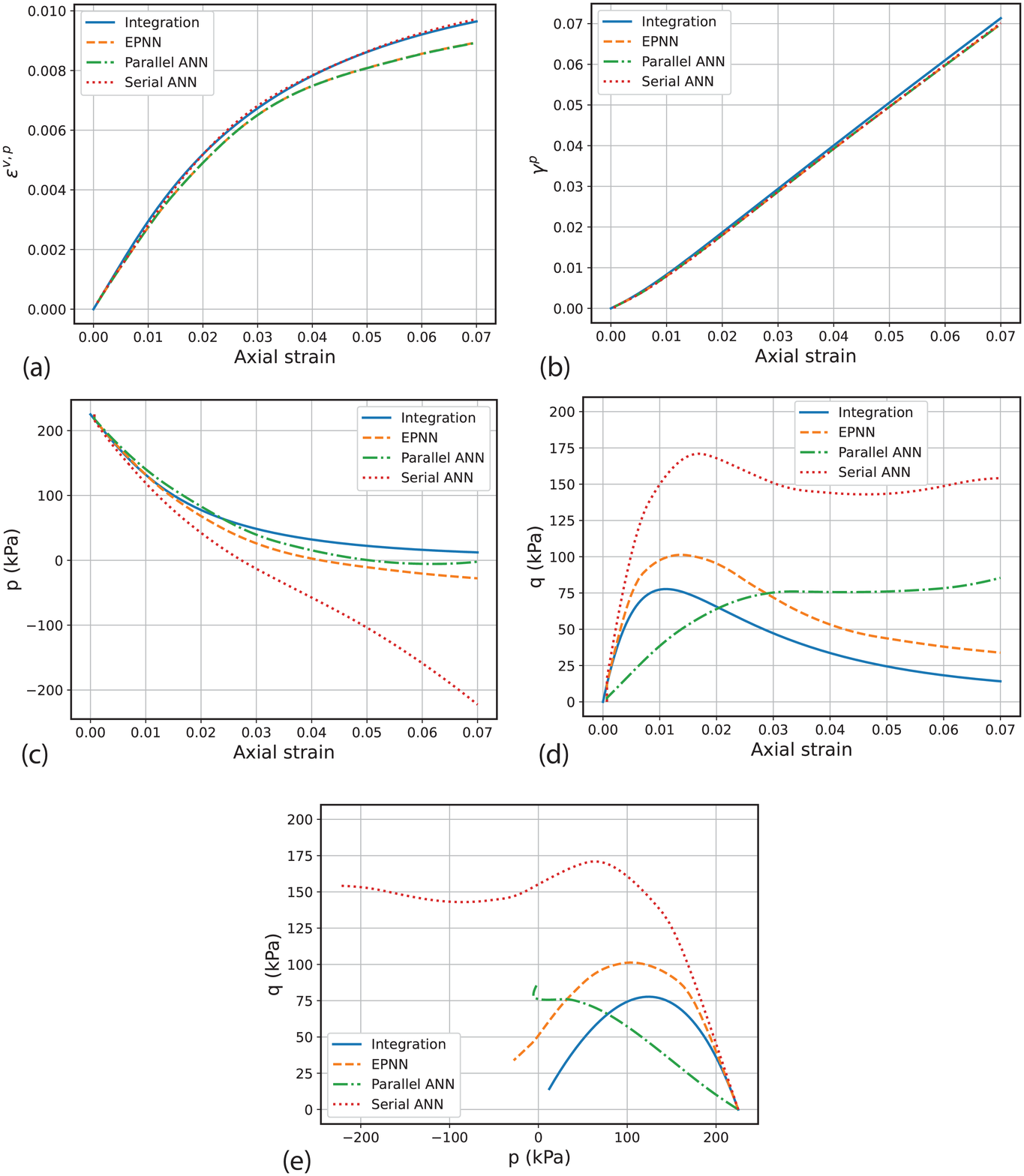}
		\caption{\label{fig:20} Comparison between the predictions of the three network architectures and the ground truth (numerical integration results) for (a) volumetric plastic strain, (b) equivalent plastic shear strain, (c) mean stress, (d) equivalent shear stress, and (e) stress path, in the undrained test with $p^{\text{in}}=225$ kPa and $e^{\text{in}}=0.72$.}
	\end{center}
\end{figure}

Further verifications on complete loading paths are provided in undrained tests with initial confining stress of $p^{\text{in}}=225$ kPa and void ratios of $0.55$, $0.62$ and $0.72$. Figures \ref{fig:21}-\ref{fig:20} show the comparison between the predictions of the three architectures versus the the ground truth. These initial void ratios are chosen to assess the predictive capabilities of neural networks for a range of soil behaviors from very loose to very dense. The strain step sizes used for all three networks in recall mode are the same and equal to the average of the random strain magnitudes used for generating the training data. The maximum equivalent shear strain applied to the material in the current tests is $\gamma=0.07$.

Based on Figure \ref{fig:21} and part (a) and (b) of Figs. \ref{fig:18}-\ref{fig:20}, all three networks provide excellent predictions on the evolution of plastic strain and void ratio during the test. In particular, the void ratio sub-network correctly predicts no change in void ratio during the undrained test where the volumetric strain is equal to zero; the EPNN however offers significant improvement in stress path predictions as can be see in part (e) of Figs. \ref{fig:18}-\ref{fig:20}, due to its physics-informed structure.

\subsection{Effect of Training Size on Prediction of EPNN}
\begin{figure}
	\begin{center}
		\includegraphics[width=1.0\linewidth]{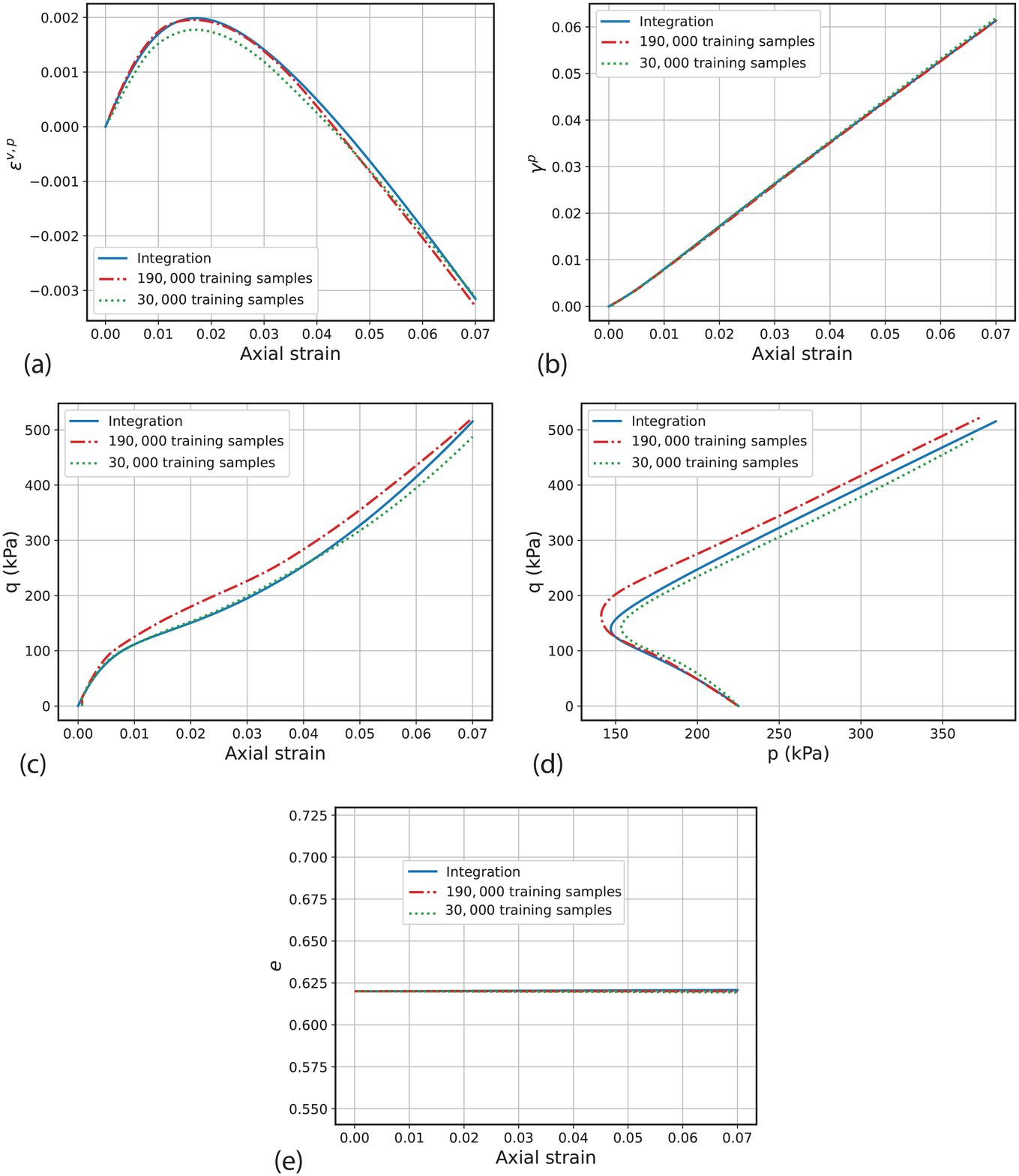}
		\caption{\label{fig:27} Comparison between the predictions of EPNNs trained with different number of training samples and the ground truth (numerical integration results) for (a) volumetric plastic strain, (b) equivalent plastic shear strain, (c) equivalent shear stress, (d) stress path, and (e) void ratio in the undrained test with $p^{\text{in}}=225$ kPa and $e^{\text{in}}=0.62$.}
	\end{center}
\end{figure}

Based on the learning curve presented in Fig. \ref{fig:15}, one can expect that the EPNN architecture can potentially be trained with a fraction of the current training data, without compromising the accuracy of the predictions. To explore this, we trained the EPNN based on approximately $30,000$ training samples ($16\%$ of the original training set size), and compared its predictions with the EPNN trained based on the full set of training data ($190,000$ training samples) in Fig. \ref{fig:27} for the case with $p^{\text{in}}=225$ kPa and $e^{\text{in}}=0.62$. It is seen that reducing the size of training samples has no significant effect on the accuracy of the EPNN predictions in this particular loading path. However, we refrain from generalizing this statement due to the need for incorporating additional physics in the plastic strain sub-network based on the results of Fig. \ref{fig:15}.

\subsection{Effect of Number of Training Epochs on Predictions of EPNN}
\begin{figure}
	\begin{center}
		\includegraphics[width=0.55\linewidth]{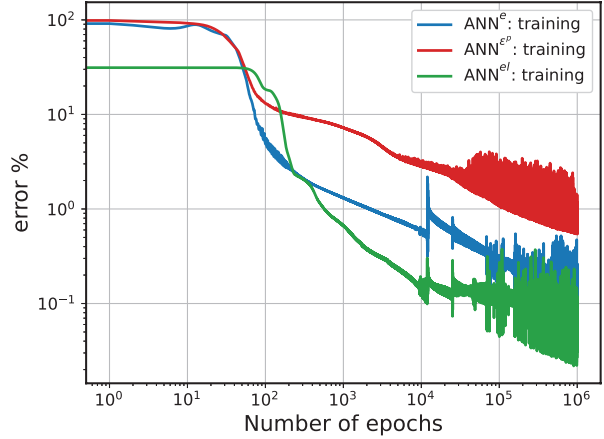}
		\caption{\label{fig:22} Training curves for the void ratio, plastic strain and stress sub-networks in the EPNN architecture trained for $1e6$ epochs.}
	\end{center}
\end{figure}

\begin{figure}
	\begin{center}
		\includegraphics[width=1.0\linewidth]{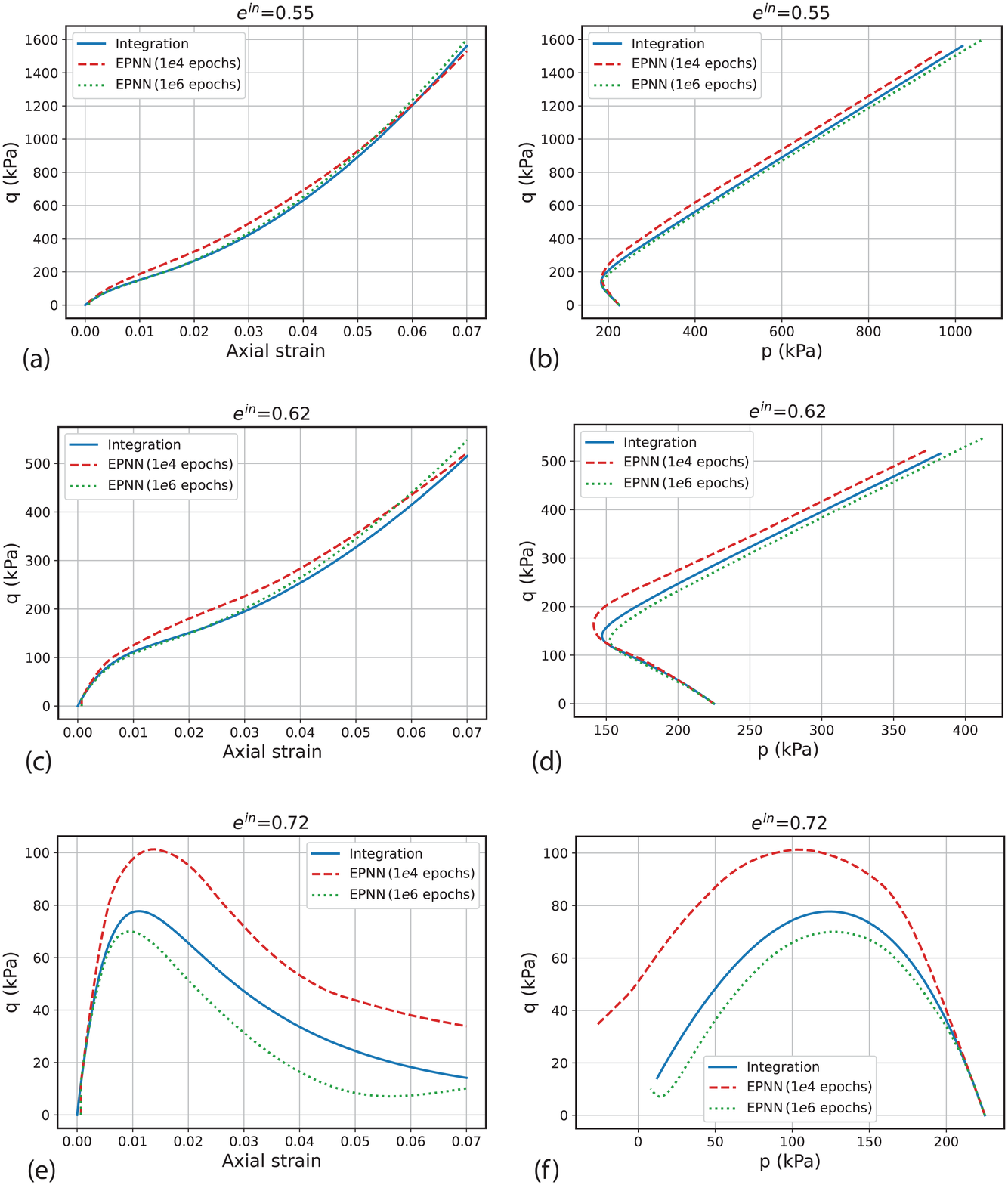}
		\caption{\label{fig:23} Comparison between the stress predictions of EPNNs trained for $1e4$ and $1e6$ epochs versus the the ground truth (numerical integration results), in the undrained tests with $p^{\text{in}}=225$ kPa and different $e^{\text{in}}$.}
	\end{center}
\end{figure}

The low training runtime achieved by using a high performance GPU allows for even further improvement of the predictions of EPNN by increasing the number of training epochs. Figure \ref{fig:22} shows the training curve for the EPNN architecture for $1e6$ epochs of training, where it is seen that a lower training error is achieved for the plastic strain sub-network after extensively training the network. The newly trained EPNN is then used in recall mode to predict the same undrained tests in Fig. \ref{fig:21}-\ref{fig:20}. Figure \ref{fig:23} shows the stress predictions of networks trained for $1e4$ and $1e6$ epochs versus the ground truth in undrained tests with different initial void ratios. In these cases, further training of EPNN has significantly improved its stress predictions.

\subsection{Effect of Loading Increment Size on Predictions of EPNN}
\begin{figure}
	\begin{center}
		\includegraphics[width=1.0\linewidth]{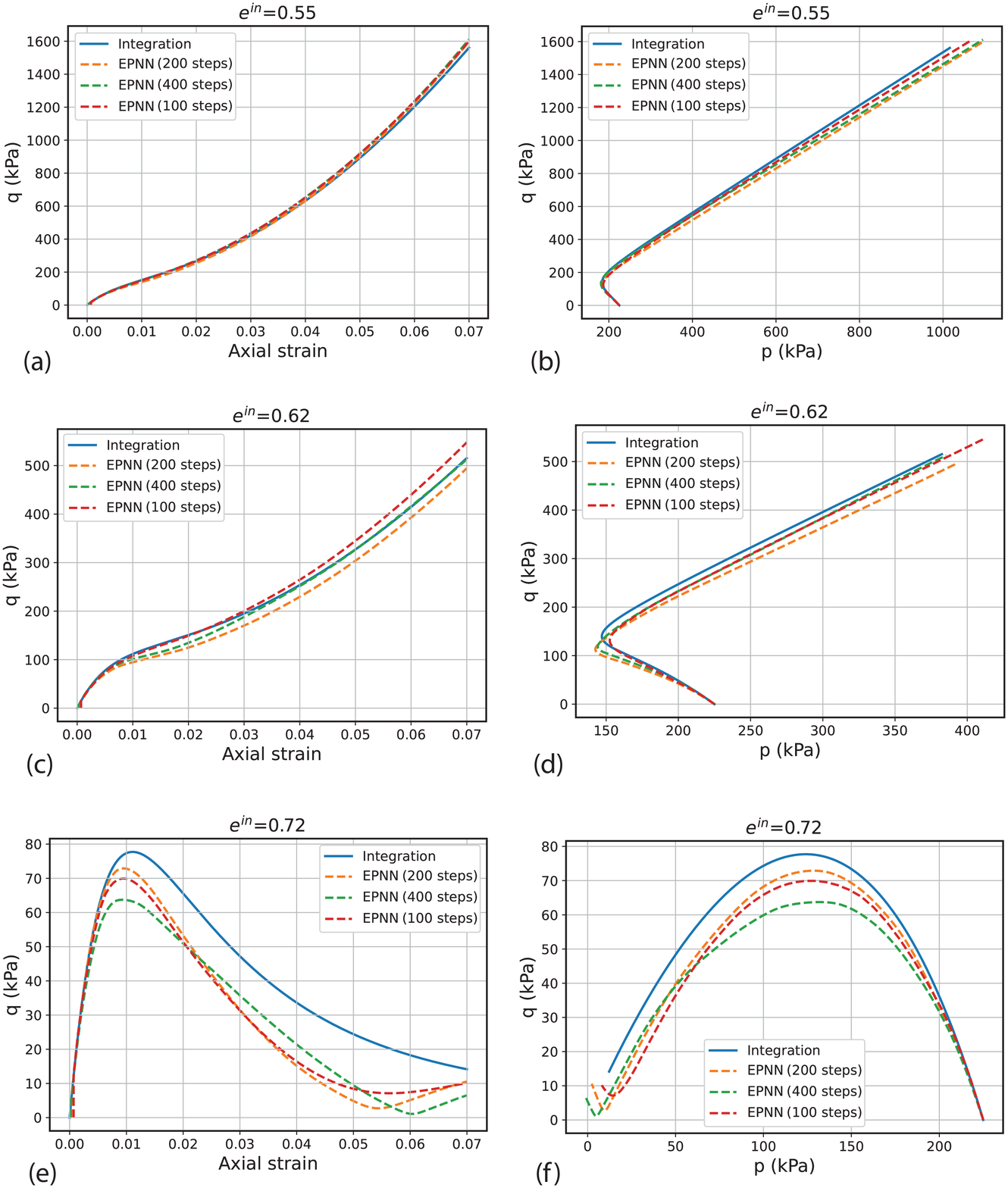}
		\caption{\label{fig:29} Predictions of EPNN (trained for $1e6$ epochs) with different strain increment sizes versus the ground truth (numerical integration results) for the undrained tests with $p^{\text{in}}=225$ kPa and different $e^{\text{in}}$.}
	\end{center}
\end{figure}

Next, we investigate the sensitivity of the EPNN's predictions to the size of the strain increment used in recall mode. The stress predictions of EPNN trained for $1e6$ epochs with different strain increment sizes are presented in Fig. \ref{fig:29} in undrained tests with $p^{\text{in}}=225$ kPa and different $e^{\text{in}}$. \citet{Lefik2003} discuss the high sensitivity of their ANN architecture to the strain increment size in recall mode. To alleviate the problem, they included auxiliary data in their training set which improved this deficiency, but still the results were not acceptable in some cases. In this work, however, as can be seen in Fig. \ref{fig:29}, the sensitivity of EPNN to the strain increment size is not severe and the results vary within an acceptable range.

\subsection{Predictions of EPNN on Axisymmetric Triaxial Paths}
\begin{figure}
	\begin{center}
		\includegraphics[width=1.0\linewidth]{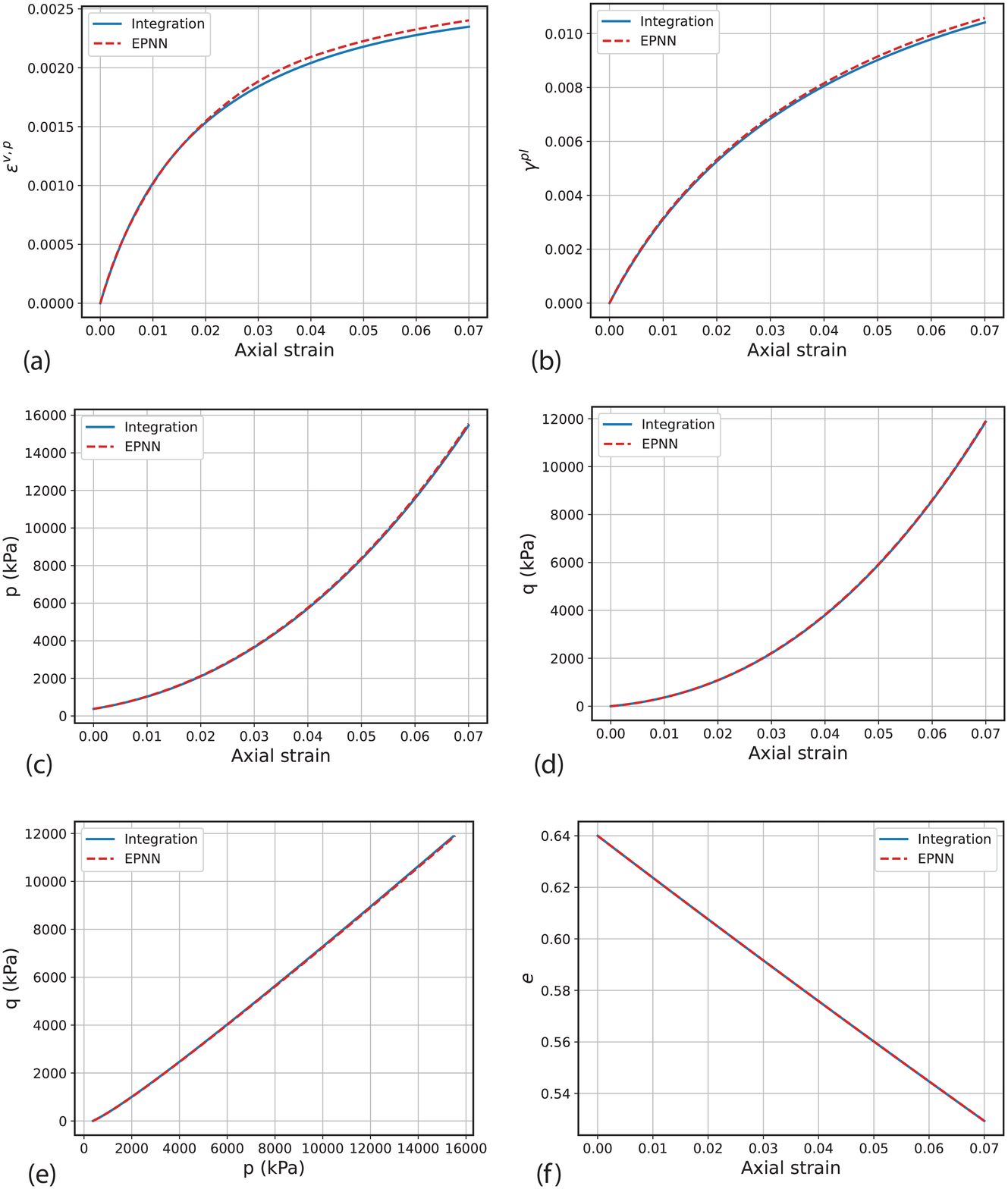}
		\caption{\label{fig:30} Comparison between the stress predictions of EPNN (trained for $1e6$ epochs) and the ground truth (numerical integration results), in the triaxial axisymmetric test with $\alpha=-1000$, $p^{\text{in}}=375$ kPa and $e^{\text{in}}=0.64$.}
	\end{center}
\end{figure}

\begin{figure}
	\begin{center}
		\includegraphics[width=1.0\linewidth]{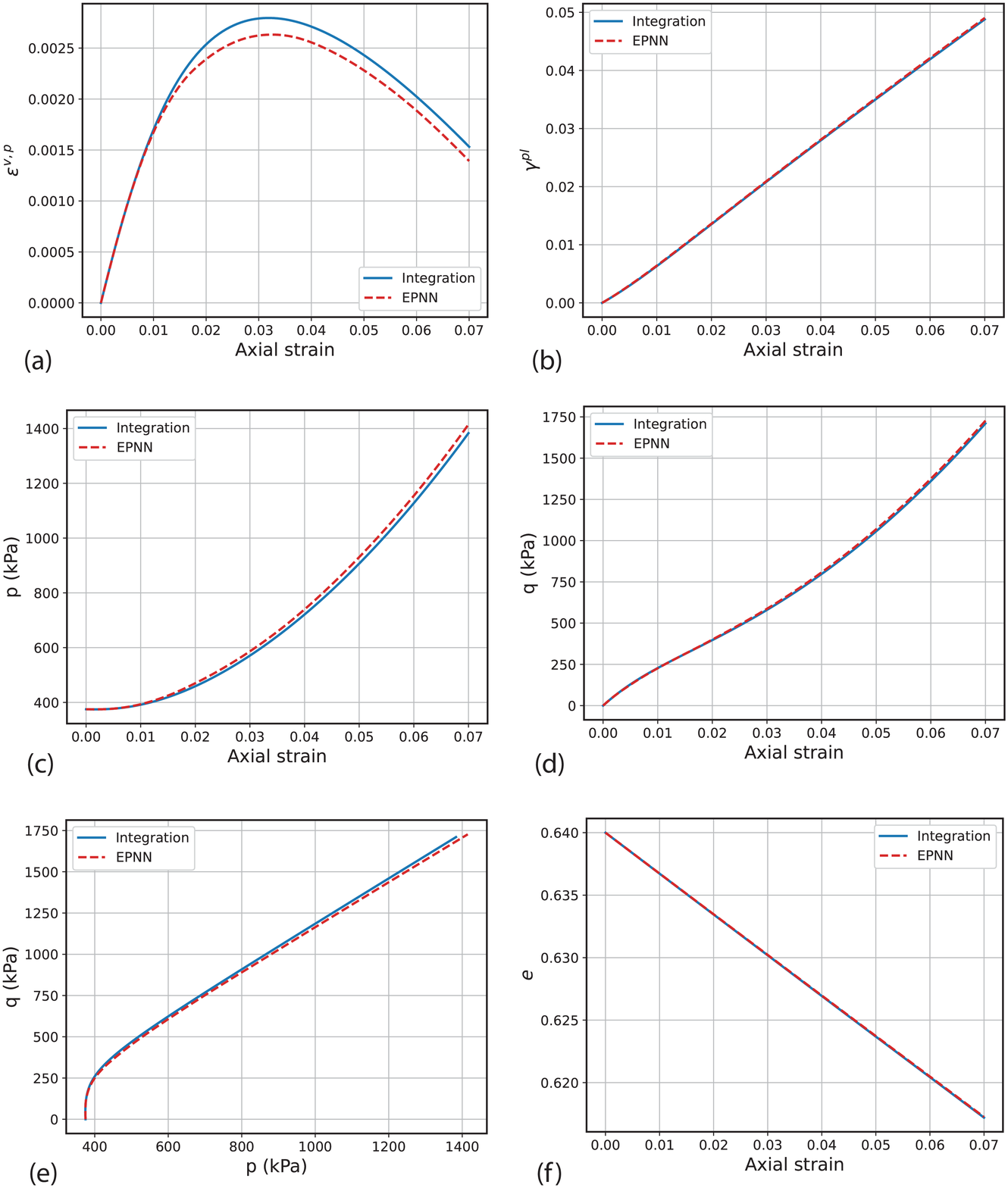}
		\caption{\label{fig:31} Comparison between the stress predictions of EPNN (trained for $1e6$ epochs) and the ground truth (numerical integration results), in the triaxial axisymmetric test with $\alpha=-2.5$, $p^{\text{in}}=375$ kPa and $e^{\text{in}}=0.64$.}
	\end{center}
\end{figure}

\begin{figure}
	\begin{center}
		\includegraphics[width=1.0\linewidth]{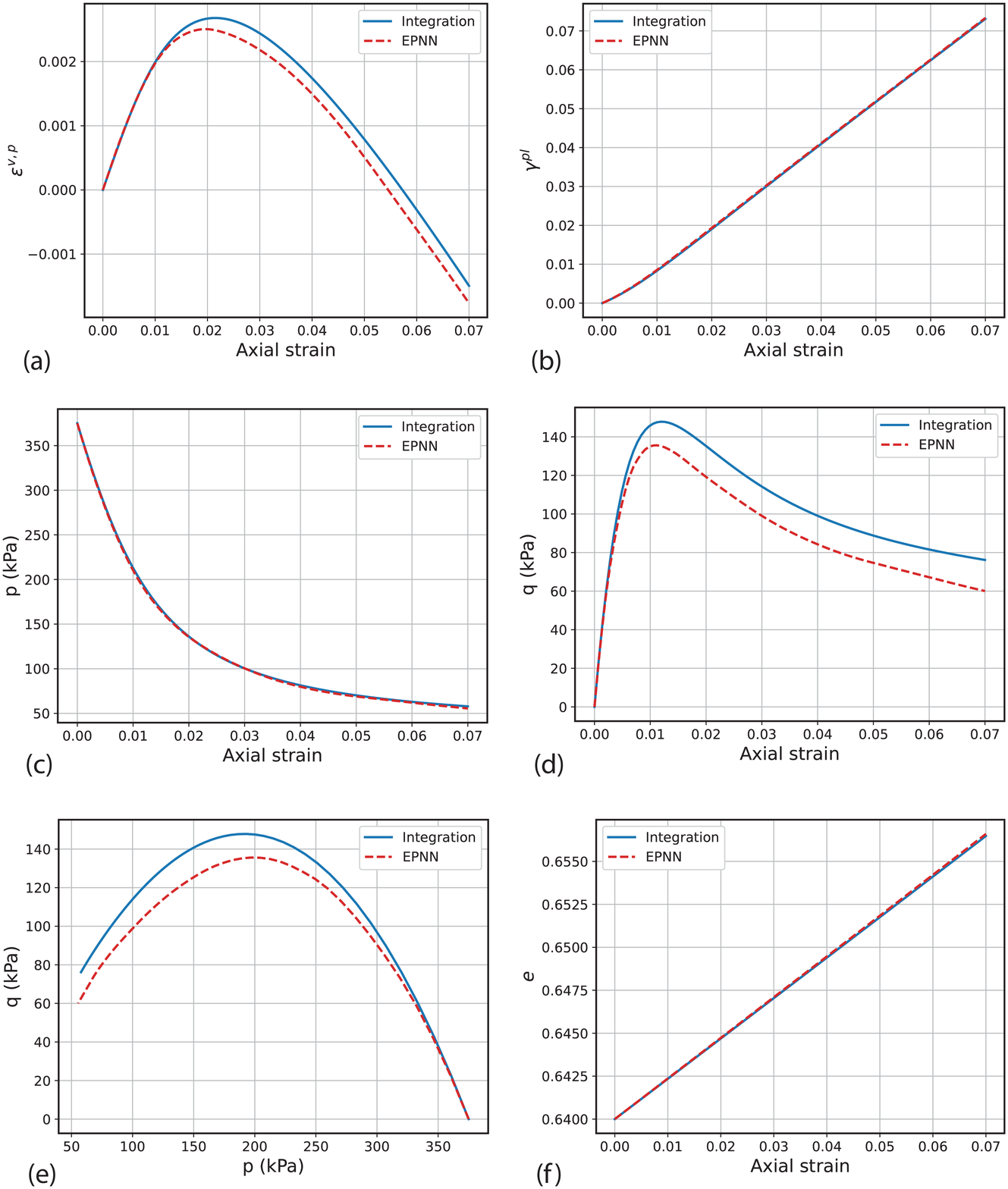}
		\caption{\label{fig:32} Comparison between the stress predictions of EPNN (trained for $1e6$ epochs) and the ground truth (numerical integration results), in the triaxial axisymmetric test with $\alpha=-1.75$, $p^{\text{in}}=375$ kPa and $e^{\text{in}}=0.64$.}
	\end{center}
\end{figure}

Finally, we consider additional loading paths to further investigate the predictive capabilities of EPNN. The EPNN that is trained for $1e6$ epochs is used to predict general axisymmetric triaxial loading paths where the strain increment in two principal directions are equal $\varepsilon_{11}=\varepsilon_{22}=\epsilon$, and the strain increment in the third (axial) direction is related to them by a proportionality factor $\alpha$, i.e. $\varepsilon_{33}=\alpha \epsilon$. The volumetric strain on the sample is thus $\varepsilon^v=(\alpha + 2)\epsilon$, while the equivalent shear strain is $\gamma=(2/3)\lvert \left(1-\alpha\right) \epsilon \rvert$. The undrained loading paths presented in the previous parts correspond to the case with $\alpha=-2$. The strain path followed in each test is linear and the material is loaded up to the axial strain of $\varepsilon_{33}=\alpha \epsilon=0.07$ in each test. The EPNN's prediction and the ground truth are compared in Figs. \ref{fig:30}-\ref{fig:32} for three different values of $\alpha$, but the same initial condition of $p^{\text{in}}=375$ kPa and $e^{\text{in}}=0.64$. The results further validate the predictive capabilities of the proposed EPNN in this work in arbitrary monotonic loading paths.

\subsection{Discussion on Further Improving the Predictions of EPNN}
Based on the verification results presented in this section, further improvement of the predictions of EPNN can potentially be achieved by augmenting the dataset used in this work with additional data obtained from a so-called incremental probing exercise. This will involve performing multiple single-step strain loadings along different strain directions at the end of each increment of loading along the proportional strain paths. The information at the beginning and end of the probing step is used as input-label pairs to augment the current dataset. Probing data is believed to provide more enriched information about the characteristics of the elasto-plastic behavior, as it explores more accurately the different tensorial zones of response. However, this is left to a separate study, as enriching the dataset makes the training of the plastic strain sub-network, which has a simple structure in this work, even more difficult; thus, as mentioned before, adding more physics to the plastic strain part of the network becomes necessary.

\section{Conclusions and Outlook}\label{sec:conclusions-and-outlook}
We presented a novel deep Physics-Informed Neural Network (PINN) architecture, named here as Elasto-Plastic Neural Network (EPNN), as surrogate constitutive model for elasto-plastic materials. The nonlinear incremental elasticity and strain decomposition are hardwired as physics into the architecture of the network and the cost function. This results in the EPNN providing more stable predictions and having superior capability in predicting unseen loading scenarios over regular ANNs with less data. The general structure of the network was adapted to plasticity of sands, and superior performance of the EPNN architecture was demonstrated in comparing its predictions with the predictions of two regular ANN architectures, namely parallel and serial ANNs, in several examples where the networks are used in recall mode to predict complete strain-controlled loading paths. Unlike many other Neural Network-based surrogate constitutive models in the literature, the proposed EPNN architecture is model and material independent. Additionally, although synthetic data was used as ground truth in the current study, the model is structured to be trained on data accessible in experimental settings. The EPNN architecture provides an incremental formulation, thus it can be implemented in standard Finite Element (FE) codes. The predictions of EPNN can be further improved by using a higher quality training data that includes strain probing at each state (sample) of the current training data. Moreover, future works should be focused on adding additional physics, such as existence of a plastic potential and flow rule, to the plastic strain sub-network.

\section*{Acknowledgment}
This work was developed within a discovery grant provided by the Natural Sciences and Engineering Research Council of Canada (NSERC). Their funding is gratefully acknowledged.


\bibliographystyle{elsarticle-harv}
\bibliography{PINN_elasto_plasticity}

\appendix

\section{WG Elasto-plasticity Model for Granular Materials}\label{appendix:1}
The elasticity in the WG model is nonlinear and follows hypo-elasticity (same form as in Eq. \eqref{eq:1}) with the isotropic tangent elasticity tensor defined as:
\begin{equation}
	C_{ijkl} = G \left(\left(R-\frac{2}{3}\right)\delta_{ij}\delta_{kl}+\left(\delta_{ik}\delta_{jl}+\delta_{il}\delta_{jk}\right)\right),
\end{equation}
where $G$ is the tangent shear modulus defined a function of stress and void ratio $e$:
\begin{equation}\label{eq:7}
	G = G^0 \frac{2.17-e}{1+e}\sqrt{p ~p^0},
\end{equation}
and $R$ is the ratio between tangent bulk modulus and tangent shear modulus:
\begin{equation}\label{eq:8}
	R = \frac{2\left(1+\nu\right)}{3\left(1-2\nu\right)}.
\end{equation}

In Eqs. \eqref{eq:7} and \eqref{eq:8}, $G^0$ is the reference shear modulus, $p=\sigma_{ii}/3$ is the mean stress, $p^0$ is the reference mean stress equal to $1$ kPa, and $\nu$ is the Poisson's ratio.

The plasticity in the WG model is based upon Rowe's stress-dilatancy relation and the critical state theory. The set of yield function $F$, plastic potential $P$ and the hardening relations in the WG model include:
\begin{subequations}
	\begin{alignat}{1}
		&F = q - M p, \label{eq:9}\\
		&P = q - N p, \\
		&M = \frac{2\mu}{\left(1+\mu\right)-\left(1-\mu\right)t} M^{\text{tc}}, \label{eq:10}\\
		&N = \frac{\sin{\varphi} - \left(\frac{e}{e^{\text{cs}}}\right)^\upalpha \sin{\varphi^{\text{cs}}}}{1-\left(\frac{e}{e^{\text{cs}}}\right)^\upalpha \sin{\varphi^{\text{cs}}}\sin{\varphi}}, \label{eq:11}\\
		&M^{\text{tc}} = \frac{6 \sin{\varphi}}{3-\sin{\varphi}}, \\
		&\sin{\varphi} = \left(\frac{e}{e^{\text{cs}}}\right)^{-\beta} \frac{\gamma^{\text{p}}}{a + \gamma^{\text{p}}} \sin{\varphi^{\text{cs}}}, \label{eq:13}\\
		&e^{\text{cs}} = e^{\text{cs},0} \exp{\left(-\left(\frac{p}{h}\right)^n\right)}, \label{eq:14}\\
		&\dot{e} = -\left(1+e\right) \dot{\varepsilon}^{\text{v}}. \label{eq:15}
	\end{alignat}
\end{subequations}

The shear yield function in the WG model follows the general Mohr-Coulomb expression (Eq. \eqref{eq:9}) where the equivalent shear stress (von Mises stress) is $q$ defined as:
\begin{equation}
	q = \sqrt{\frac{3}{2} ~s_{ij} ~s_{ij}},
\end{equation}
with $s_{ij}=\sigma_{ij}-p\delta_{ij}$ being the deviatoric stress. Equation \eqref{eq:10} is a simplified interpolating function giving the dependency of the yield function through the parameter $t$:
\begin{equation}
	t = \frac{J_3}{2} \left(\frac{3}{J_2}\right)^{3/2},
\end{equation}
where $J_2=q^2/3$ and $J_3=\det{(s_{ij})}$ are the second and third invariants of the deviatoric stress. $\mu$ in Eq. \eqref{eq:10} is a constant determining the ratio between the friction coefficient $M$ along triaxial and compression ($M^{\text{tc}}$) directions. $M^{\text{tc}}$ is defined in terms of the mobilized friction angle $\varphi$.

The plastic deformations in the WG model are governed by a non-associated flow rule to take into account the dilatancy. In Eq. \eqref{eq:11}, $e^{\text{cs}}$ and $\phi^{\text{cs}}$ are the void ratio and friction angle at critical state, respectively. Also, $N = \sin{\psi}$ is the dilatancy coefficient in terms of the dilatancy angle $\psi$.

Equations \eqref{eq:13}-\eqref{eq:15} provide the hardening laws for describing the evolution of the yield surface and plastic potential in the stress space. $\gamma^{\text{p}}=\sqrt{\frac{2}{3} ~\kappa^{\text{p}}_{ij} ~\kappa^{\text{p}}_{ij}}$ is the equivalent plastic shear strain defined based on the deviatoric plastic strain $\kappa^{\text{p}}_{ij}=\varepsilon^{\text{p}}_{ij} - \left(\varepsilon^{\text{v},\text{p}}/3\right)\delta_{ij}$. $\varepsilon^{\text{v},\text{p}}=\varepsilon^{\text{p}}_{ii}$ is the volumetric plastic strain.

\end{document}